\documentclass[journal, twoside]{IEEEtran}
\usepackage[utf8]{inputenc}
\usepackage{xcolor}

\usepackage{graphicx}
\usepackage{amsmath}
\usepackage{amssymb}
\usepackage{pifont}
\usepackage{lscape}
\usepackage{multirow}
\usepackage{tabularx,booktabs}
\usepackage[normalem]{ulem}
\usepackage[hyphens]{url}
\usepackage[hidelinks]{hyperref}
\usepackage[ruled,vlined]{algorithm2e}
\ifCLASSINFOpdf
\else
\fi
\begin{document}
%

\title{A Comprehensive Survey on Hardware-Aware \\ Neural Architecture Search}

%
%
%

\author{Hadjer Benmeziane, Kaoutar El Maghraoui, \IEEEmembership{IEEE Member}, Hamza Ouarnoughi, Smail Niar, \IEEEmembership{IEEE Senior Member},  Martin Wistuba, Naigang Wang, \IEEEmembership{IEEE Member} \newline
\IEEEmembership{} \\
This work has been submitted to the IEEE for possible publication. Copyright may be transferred without notice, after which this version may no longer be accessible.
\thanks{H. Benmeziane and H. Ouarnoughi and S. Niar are with Universit\'{e} Polytechnique Hauts-de-France, LAMIH/CNRS,Valenciennes, France. E-mail: (firstname.lastname@uphf.fr)}
\thanks{K. El Maghraoui and  Naigang Wang are with IBM T. J. Watson Research Center, Yorktown Heights, NY 10598, USA. Email: (kelmaghr@us.ibm.com, nwang@us.ibm.com) }
\thanks{Martin Wistuba is with IBM Research AI, IBM Technology Campus, Dublin, Ireland. Email: (martin.wistuba@ibm.com)}}

%
%


\markboth{Benmeziane \MakeLowercase{\textit{et al.}}: A Comprehensive Survey on Hardware-aware Neural Architecture Search}{Benmeziane \MakeLowercase{\textit{et al.}}: A Comprehensive Survey on Hardware-aware Neural Architecture Search}
%



\maketitle
\begin{abstract}Neural Architecture Search (NAS) methods have been growing in popularity. 
These techniques have been fundamental to automate and speed up the time consuming and error-prone process of synthesizing novel Deep Learning (DL) architectures. 
NAS has been extensively studied in the past few years. Arguably their most significant impact has been in image classification and object detection tasks where the state of the art results have been obtained. 
Despite the significant success achieved to date, applying NAS to real-world problems still poses significant challenges and is not widely practical. 
In general, the synthesized Convolution Neural Network (CNN) architectures are too complex to be deployed in resource-limited platforms, such as IoT, mobile, and embedded systems. 
One solution growing in popularity is to use multi-objective optimization algorithms in the NAS search strategy by taking into account execution latency, energy consumption, memory footprint, etc. 
This kind of NAS, called hardware-aware NAS (HW-NAS), makes searching the most efficient architecture more complicated and opens several questions. 
In this survey, we provide a detailed review of existing HW-NAS research and categorize them according to four key dimensions: the search space, the search strategy, the acceleration technique, and the hardware cost estimation strategies. We further discuss the challenges and limitations of existing approaches and potential future directions. This is the first survey paper focusing on hardware-aware NAS. We hope it serves as a valuable reference for the various techniques and algorithms discussed and paves the road for future research towards hardware-aware NAS.
\end{abstract}


%
\IEEEpeerreviewmaketitle

\section{Introduction}
\label{Section:introduction}
%
%
%
%
\IEEEPARstart{D}{eep} Learning (DL) systems are revolutionizing technology around us across many domains such as computer vision \cite{sermanet2013overfeat, simonyan2014very, zeiler2014visualizing, krizhevsky2012imagenet}, speech processing \cite{park2020improved, microsoft, hinton2012deep} and Natural Language Processing (NLP) \cite{devlin2018bert, collobert2011natural, wu2016google}. 
These breakthroughs would not have been possible without the availability of big data, the tremendous growth in computational power, advances in hardware acceleration, and the recent algorithmic advancements. 
However, designing accurate neural networks is challenging due to: 
\begin{itemize}
    \item The variety of data types and tasks that require different neural architectural designs and optimizations. 
    \item The vast amount of hardware platforms which makes it difficult to design one globally efficient architecture.  
\end{itemize}

For instance, certain problems require task-specific models, e.g. EfficientNet \cite{tan2019efficientnet} for image classification and ResNest \cite{zhang2020resnest} for semantic segmentation, instance segmentation and object detection. 
These networks differ on the proper configuration of their architectures and their hyperparameters. The hyperparameters here refer to the pre-defined properties related to the architecture or the training algorithm.

\begin{figure}[!ht]
    \centering
    \includegraphics[width=0.5\textwidth]{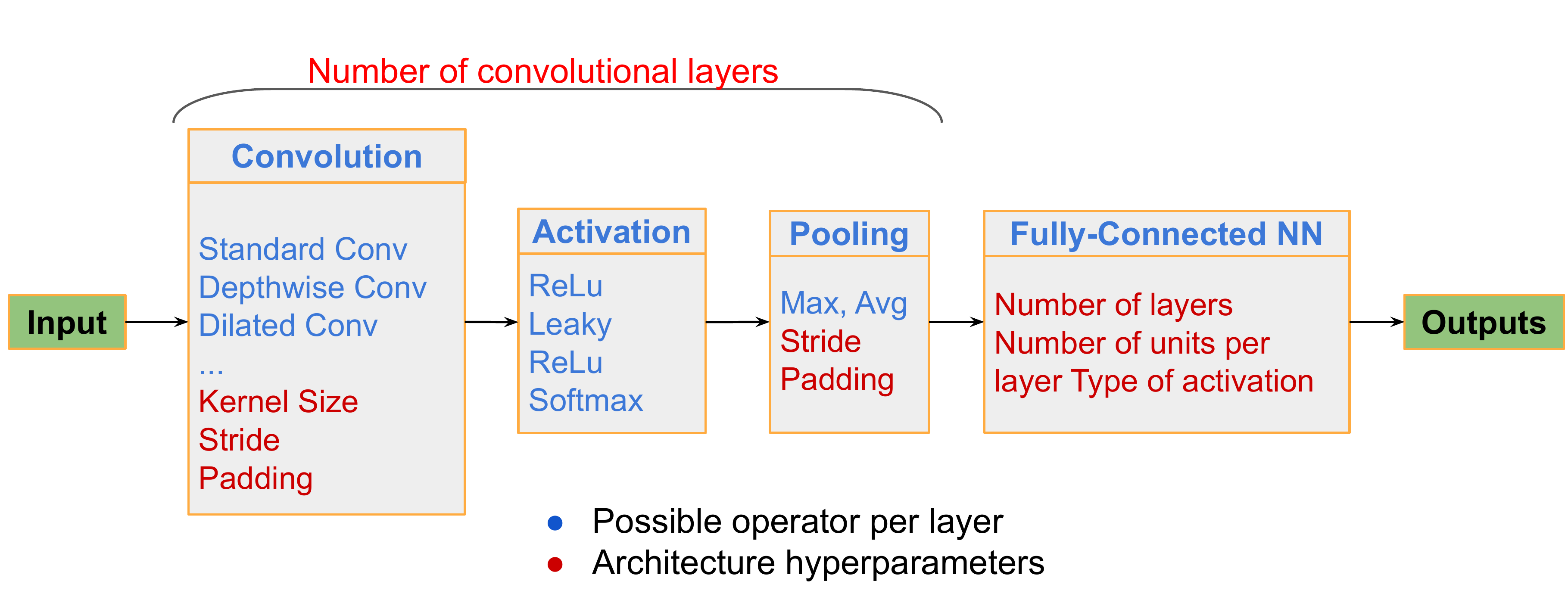}
    \caption{Generic CNN architecture. For each layer an operator is chosen among a pre-defined list (convolution, dilated convolution, depthwise convolution, maxpooling, batch\_normalization...)}
    \label{fig:general_nn}
\end{figure}

In general, the neural network architecture can be formalized as a Directed Acyclic Graph (DAG) where each node corresponds to an operator applied to the set of its parent nodes~\cite{Goodfellow-et-al-2016}. Example operators are convolution, pooling, activation, and self-attention. Linking these operators together gives rise to different architectures. A key aspect of designing a well-performing deep neural network is deciding the type and number of nodes and how to compose and link them. Additionally, the architectural hyperparameters (e.g., stride and channel number in a convolution, etc.) and the training hyperparameters (e.g., learning rate, number of epochs, momentum, etc.) are also important contributors to the overall performance. Figure \ref{fig:general_nn} shows an illustration of some architectural choices for the type of convolutional neural network.

According to this representation, DL models can contain hundreds of layers and millions or even billions of parameters. These models are either handcrafted by repetitive experimentation or modified from a handful of existing models. These models have also been growing in size and complexity. All of this renders handcrafting deep neural networks a complex task that is time-consuming, error-prone and requires deep expertise and mathematical intuitions. 
Thus, in recent years, it is not surprising that techniques to automatically design efficient architectures, or NAS for "Neural Architecture Search", for a given dataset or task, have surged in popularity. 

\begin{figure*}[!ht]
    \centering
    \includegraphics[width=16cm]{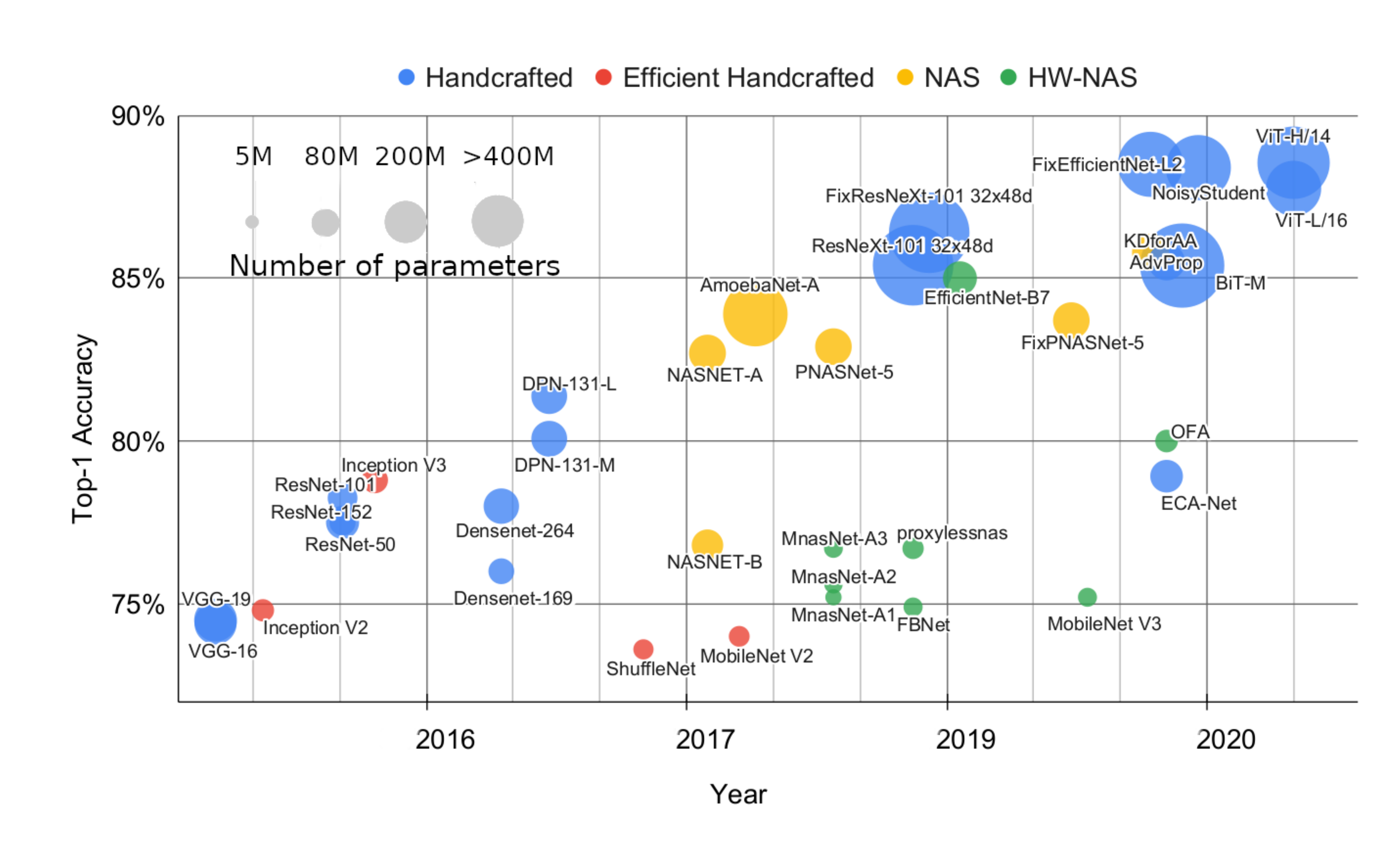}
    \caption{Accuracy of various CNN models on ImageNet for Image Classification task with the number of parameters. Inspired by \cite{9043731} }
    \label{fig:complexity_model}
\end{figure*}

In figure \ref{fig:complexity_model}, we compare several deep learning models for the image classification task. Each dot in the plot corresponds to a given DL architecture that has been used for image classification. The dot size correlates with the size of the corresponding neural network in terms of the number of parameters. A quick look at the graph reveals the trend to design larger models to better Top 1 accuracy. However, the size is not necessarily correlated with better accuracy. There have been several efforts to conceive more efficient and smaller networks to achieve comparable Top 1 accuracy performance. 
We compare four classes of models: Handcrafted, Efficient handcrafted, NAS, and HW-NAS. 
Generally, throughout the years the handcrafted models rank high up in terms of accuracy but are much more complex in terms of the depth of the architecture or the large number of parameters. For instance, ViT-H \cite{anonymous2021an}, which is the state-of-the-art model as of December 2020, has over 600 million parameters and 32 layers. 
In the top right quadrant of the figure \ref{fig:complexity_model} (around the same region as most of the recently handcrafted models), we find some of the models that are automatically created by different NAS techniques. These latter techniques focus only on improving the model's accuracy without paying attention to the efficiency of the model in terms of its size and latency.  Therefore, these NAS models are still large, with the number of parameters ranging between 100M and 500M. 

Since 2015, we have noticed the rise of efficient handcrafted models. These models rely on compression methods (see section \ref{section:compression}) to decrease the model's size while trying to maintain the same accuracy. 
MobileNet-V2 \cite{mobilenet_v2} and Inception \cite{inception_v3} are good examples where the number of parameters is between 20M and 80M. 
This paper focuses on the class of hardware or platform-aware NAS techniques: \textit{HW-NAS}. 
This class encompasses work that aims to tweak NAS algorithms and adapt them to find efficient DL models optimized for a target hardware device. HW-NAS began to appear in 2017 and since then achieved state of the art (SOTA) results in resource-constrained environments with Once-for-all (OFA) \cite{cai2020onceforall} for example.

\begin{figure}[!ht]
    \centering
    \includegraphics[width=0.5\textwidth]{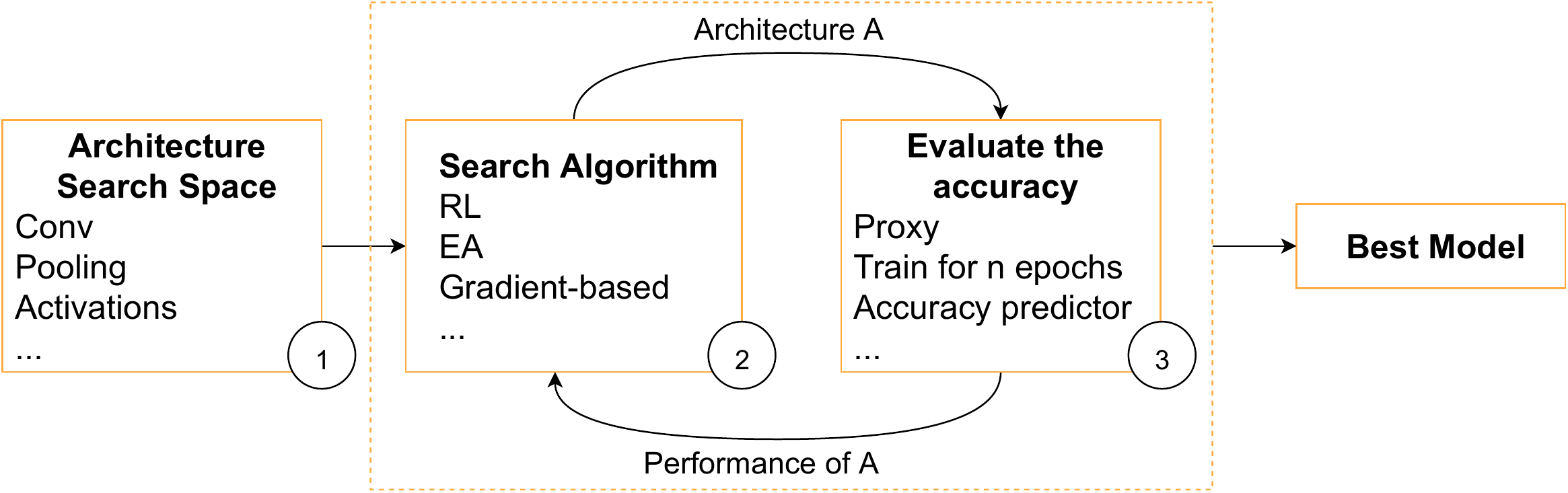}
    \caption{Overview of conventional NAS components}
    \label{fig:nas}
\end{figure}

A conventional NAS process requires the definition of three main components: the search space, the search strategy, and the evaluation methodology, as illustrated in figure \ref{fig:nas}. First, the {\it{search space}} (1) draws from a set of neural network architectures to define the neural network operators and how they are connected to form a valid network. It determines how the architectures can be formed and what architectures are allowed. For example, NASNet \cite{nasnet} introduced a fixed macro architecture where the search consists of finding the appropriate operators used within each block from a set of 12 specified operators. This search space is explored by a {\it{search strategy}} (2) which samples a population of network architectures' candidates. It evaluates the accuracy of the model using a specific {\it{evaluation methodology}} (3). The measured accuracy will then guide the search strategy to converge towards promising architectures in the search space. Generally, the evaluation component will train the architecture on the desired dataset, which often takes a considerable time. Many NAS algorithms have incorporated several techniques to speed up the training process described in Section \ref{section:speedup}. 

NAS has proven its efficiency by proposing several SOTA models in Object Detection \cite{ghiasi2019nasfpn} and Image Classification \cite{howard2019searching}.  However, these models are often composed of millions of parameters and require billions of floating-point operations (FLOP). This causes a large memory footprint and big FLOP preventing their usage in resource-constrained environments. Additionally, these models might require specific hardware (e.g. GPUs, TPUs, etc.) to allow their deployment in a reasonable time or real-time applications. 

Since the inception of the very first NAS algorithm with reinforcement learning \cite{zoph2017neural}, there has been tremendous work to study efficient NAS. More recently, integrating hardware awareness in the search loop (i.e. HW-NAS) has attracted several researchers and has opened up interesting new research directions. Some HW-NAS efforts have demonstrated SOTA results and have balanced the trade-off between accuracy and hardware efficiency. For example, FBNet \cite{wu2019fbnet} has achieved SOTA results on ImageNet by using an objective function that minimizes both the cross-entropy error that leads to better accuracy and the latency that penalizes networks that are too slow.

This paper provides a detailed overview of existing HW-NAS research efforts and categorizes them according to their goals and problem formulation. There has been no survey dedicated to hardware-aware NAS in the literature to the best of our knowledge.
With this survey, we provide a concise review of the NAS variants that focus on precision and hardware awareness. We hope that it will be beneficial to those with a basic understanding of NAS interested in (i) getting a comprehensible but accessible overview of how the multi-objective problem is solved and  (ii) a succinct reference and guide of existing HW-NAS techniques. 

Our review is divided into several sections. In section {\ref{Section:previous_surveys}}, we highlight the relevant NAS surveys and explain the focus and importance of our survey compared to them. In section {\ref{Section:Background}}, we present a brief overview of the efficient DL methods and where HW-NAS is situated among them.  We also explain some of the concepts needed to understand the components of the NAS process. In section \ref{section:taxonomy}, we classify the different HW-NAS works based on their goals and target platforms. Based on the taxonomy, we introduce and define two search spaces: Architecture Search Space and Hardware Search Space in section \ref{Section:Search_Space}. In section \ref{Section:HW_NAS_Formulation}, we formally define the HW-NAS optimization problem and how the multi-objective function is formulated. The following section \textit{Search Strategies} is divided into three subsections. We first discuss, in section \ref{section:sa}, the various algorithms used to explore the search space, including reinforcement learning and evolutionary algorithms. In section \ref{section:non_diff}, we give insights on how NAS algorithms take care of the non-differentiable variables. And in section \ref{section:speedup}, we list the methods to speed up the search process and avoid training each sampled architecture. We dedicate section \ref{section:hw_cost} to the study of the hardware metrics used by HW-NAS and the methods used to measure them. In section \ref{Section:Other_Considerations}, we introduce other considerations that either tackle the hardware efficiency objective, model compressions such as automatic quantization and pruning or add other objectives to the NAS like the robustness against adversarial attacks. We discuss the industrial adaptation of NAS by comparing several tools proposed by the commercial and research communities. 
We conclude in sections {\ref{Section:Challenges_Limitations}} and {\ref{Section:Conclusion}} by highlighting the challenges and limitations of current HW-NAS and discuss possible future directions. We summarized the referenced body of work in the table: \url{https://tinyurl.com/y6458skt}.

\begin{table*}[!t]
    \centering
    \begin{tabular}{c|c c c}
        \hline
        \hline 
         \multirow{2}{*}{Background} & \multirow{3}{*}{Efficient Deep Learning} & Model Compression & Section 
         \ref{section:compression}
         \\ 
         && HW-NAS & Section \ref{section:back_hw_nas} \\
         && Code Transformations & Section \ref{section:code_transformation} \\ \cline{2-4}
         & \multirow{2}{*}{Algorithms} & Reinforcement Learning & Section \ref{section:back_rl} \\
         && Evolutionary Algorithm & Section \ref{section:back_ea} \\ \hline \hline
         Taxonomy of HW-NAS & \multicolumn{2}{c}{Classification of HW-NAS based on their Goals}  & Section \ref{section:taxonomy}\\
         \hline 
         \multirow{2}{*}{Search Spaces} & \multicolumn{2}{c}{Architecture Search Space} & Section \ref{section:arch_ss} \\ \cline{2-4}
         & \multicolumn{2}{c}{Hardware Search Space} & Section \ref{section:hss}\\ \hline
         \multirow{2}{*}{HW-NAS Problem Formulation} & \multirow{2}{*}{Single-Objective Optimization} & Two-Stage Search & Section \ref{section:two-stage} \\ 
         && Constrained Optimization & Section \ref{section:constrained} \\ \cline{2-4} 
         &\multirow{2}{*}{Multi-Objective Optimization} & Scalarization & Section \ref{section:scalarization} \\
         && NSGA-II & Section \ref{section:nsga}\\
         \hline
         \multirow{3}{*}{Search Strategies} & \multirow{4}{*}{Search Algorithm} & Reinforcement Learning & Section \ref{section:rl} \\ 
         && Evolutionary Algorithm & Section \ref{section:ea}\\ 
         && Gradient-based Methods & Section \ref{section:gradient} \\ 
         && Bayesian Optimization \& Random Search & Section \ref{section:rs_bo} \\ \cline{2-4}
         & Non Differentiable Techniques & Over-parameterized Networks \& their training & Section \ref{section:non_diff}\\
         \cline{2-4}
        & \multirow{4}{*}{Runtime Performance Optimization Strategies} 
        & Early Stopping & \multirow{3}{*}{Section \ref{section:speedup}} \\ 
          && Hot Start & \\
          && Proxy Datasets & \\
          && Accuracy Prediction Models & \\ 
          \hline 
          \multirow{4}{*}{Hardware Cost Estimation Methods} & \multirow{4}{*}{Hardware Constraints Collection Techniques} & Real-time measurements & \multirow{4}{*}{Section \ref{section:hw_cost}} \\
          && Lookup Table & \\ 
          && Analytical Estimation & \\
          && Prediction Models& \\ \hline
          
          \multirow{3}{*}{Other Considerations for HW-NAS} & \multicolumn{2}{c}{Automatic Mixed-Precision Quantization} & Section \ref{section:quantization}\\ 
          & \multicolumn{2}{c}{Automatic Pruning} & Section \ref{section:pruning} \\
          & \multicolumn{2}{c}{Security and Reliability} & Section \ref{section:security} \\ 
          \hline \hline 
          \multirow{2}{*}{Discussions}  & \multicolumn{2}{c}{Industrial Adaptation of NAS} & Section \ref{section:industry}\\ \cline{2-4}
          & \multirow{3}{*}{Challenges \& Limitations} & Benchmarking \& Reproducibility & Section \ref{section:bench} \\ 
          && Transferability over Tasks & Section \ref{section:transfer}\\
          && Transferability over HW Platforms & Section \ref{section:hw_transfer}\\  \cline{2-4}
          &  \multicolumn{2}{c}{Outlook \& Future Directions} & Section \ref{section:outlook}\\ \hline
    \end{tabular}
    \caption{Content Guidance for this Survey}
    \label{tab:content}
\end{table*}

\section{Existing Surveys on NAS}
\label{Section:previous_surveys}

\begin{figure}[!ht]
    \centering
    \includegraphics[width=9cm]{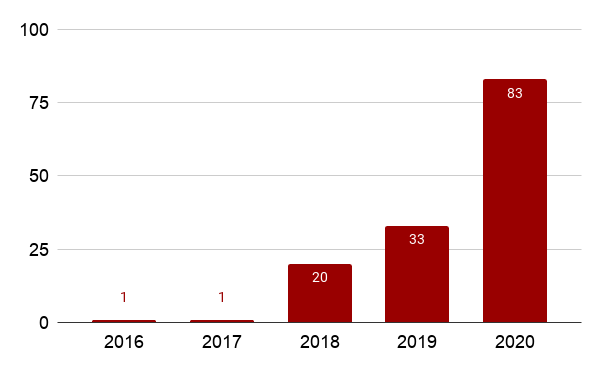}
    \caption{Number of papers describing HW-NAS \textit{by Dec 2020}. The top 5 conferences and journals are: NeurIPS, ECCV, IEEE Transactions on Pattern Analysis and Machine Intelligence, IJCNN, and MICCAI.}
    \label{fig:popularity}
\end{figure}

In the past few years, a large number of research papers have surveyed the state of the art neural architecture search algorithms. Two of the surveys we have examined stood out. The first survey, published in 2018 \cite{elsken2019neural}, describes the end-to-end design of a NAS algorithm. It details the components (i.e., Search Space, Search Strategy and Performance Estimation Strategy) and gives a clear taxonomy of the optimization methods. In 2019, a similar survey was published by Wistuba et al. \cite{wistuba2019survey}, which describes the different components of NAS algorithms with a focus on the optimization methods. It also includes a section that describes a high-level multi-objective neural architecture search and how it can be formulated and solved.
While existing NAS surveys have succeeded at giving a thorough understanding of how general NAS algorithms work and explaining the various concepts that allow NAS techniques to achieve SOTA results, they fell short in including enough details about HW-aware NAS algorithms and the emerging efforts that strive to render NAS-generated models practical and amenable to deployment in today's constrained hardware platforms. Zhang et al. \cite{FPGA} surveyed a subset of hardware-aware NAS algorithms that only focus on FPGA platforms. There is a growing effort investigating the intersection of NAS and hardware platforms as depicted in Figure \ref{fig:popularity}, where we can see the increasing number of research papers that are focusing on hardware-aware NAS. This work is a first attempt at filling the gap we have identified in existing NAS surveys. We aim to provide a detailed overview and analysis of existing HW-NAS research efforts. Table~\ref{tab:content} shows the structure of the study and serves as a guide to the various sections and their content. 

\textbf{Focus of this paper} In this survey, we provide a comprehensive overview of the architecture and hardware search spaces including how the HW-NAS problem is formulated, the search strategies, the hardware cost estimation methods, and the plethora of hardware platforms that have been targeted. We compare the used multi-objective search strategies such as reinforcement learning or evolutionary algorithm, along with some techniques that use non-differentiable parameters which allow the use of an over-parameterized network. We also study different methods that collect the hardware cost metrics used as part of search optimization function.  It is worth mentioning that this paper does not include Training Hyperparameter Search Optimization methods (THPO), and solely focuses on the operator selection and architectural hyperparameters.


\section{Background}
\label{Section:Background}
In this section, we define key terms needed to grasp the ideas surrounding HW-NAS and the concepts discussed in the following sections.

\subsection{Methodologies for Efficient Deep Learning}

\begin{figure}[!ht]
    \centering
    \includegraphics[width=8cm]{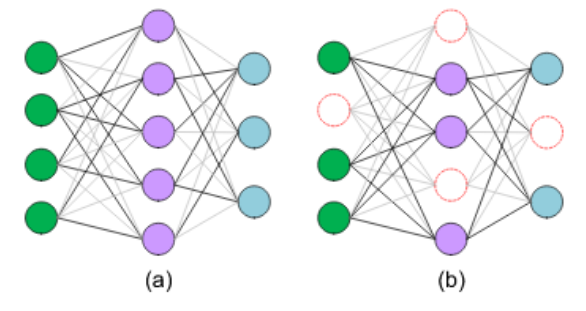}
    \caption{Illustration of sparsification. (a) Weight Pruning. (b) Neuron Pruning. Gray lines correspond to pruned vertices (i.e. weights) and white nodes correspond to pruned neurons. Source \cite{9043731}}
    \label{fig:my_label}
\end{figure}

\begin{figure*}[!ht]
        \centering
        \includegraphics[width=\textwidth]{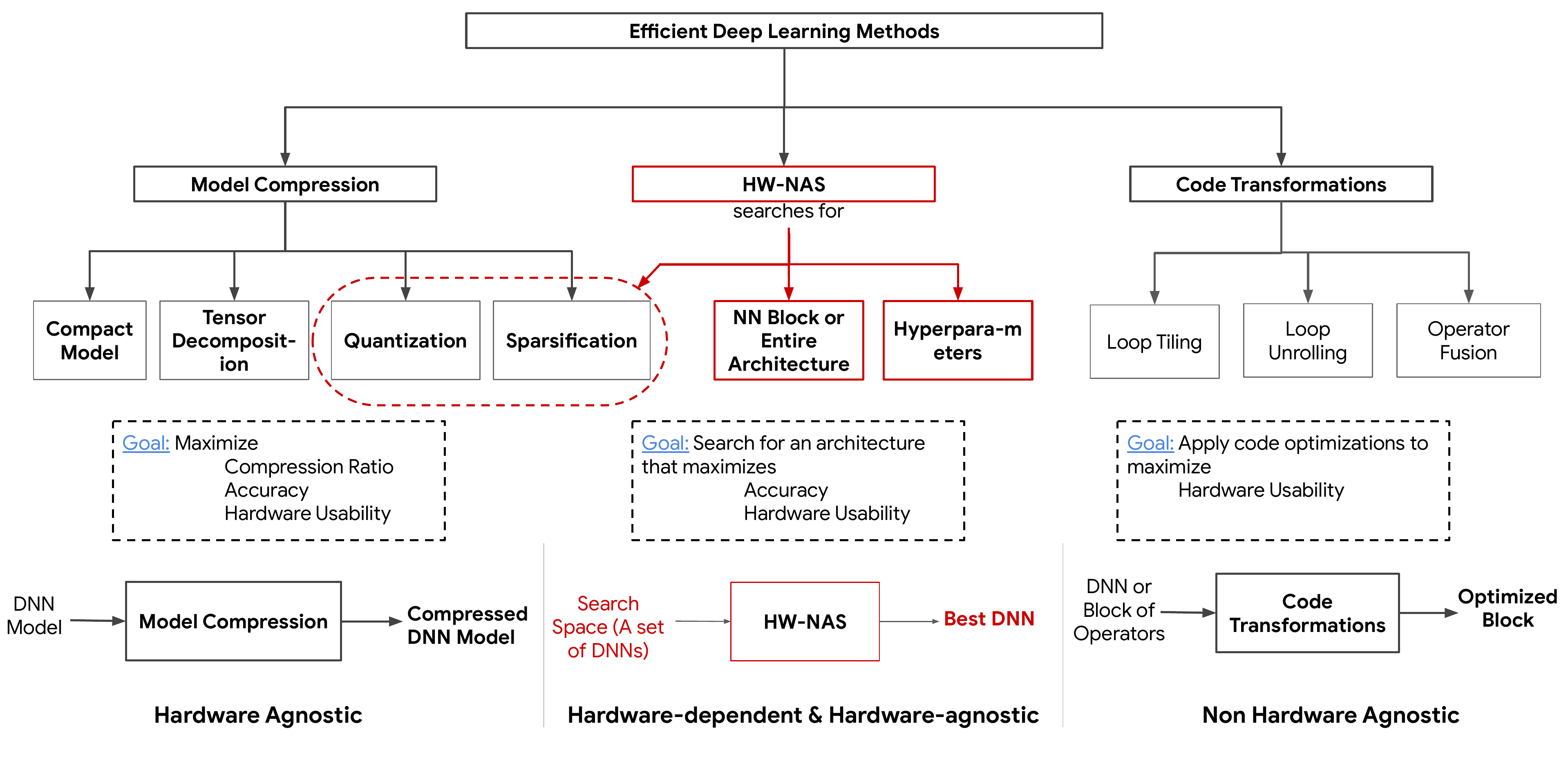}
        \caption{Overview of efficient deep learning techniques}
        \label{fig:efficient}
\end{figure*}

The significant advances and breakthroughs of deep learning, that have propelled from academic and industrial research labs’ circles, are now electrifying the computing industry and transforming the world. Deep learning is now being largely used to solve real-world problems. As deep learning is computationally demanding, most of the deployments happen in the cloud or on-premises data centers. However, with the arrival of powerful and low-energy consumption Internet of Things (IoT) devices and the growing need to take action in real or near-real-time, deep learning computations are increasingly moving to the edge. Edge devices pose several challenges in this context, as they are constrained with limited energy and computational power. For example, autonomous driving cars depend on real-time object detection of the environment. They cannot tolerate the additional latency of sending the data to the cloud, processing the data and then sending it back to the edge device.
Moreover, the compute capacity at the edge is significantly low, which does not match the increasing DL models complexity. This has motivated the research community to find innovative ways to reduce the DL models' size, their required number of floating operations, and their inference latency. This section presents an overview of the efficient deep learning techniques and where HW-NAS is situated among them. 
    
Figure \ref{fig:efficient} illustrates the taxonomy of different techniques used to optimize deep learning models.

\subsubsection{Model Compression} \label{section:compression}
These methods aim to take one standard deep learning model such as ResNet or AlexNet and apply some optimizations that will decrease the model size and the number of FLOPs, i.e., increase the compression ratio, while trying to maintain the same accuracy. Relevant surveys~\cite{9043731, model_compression} on model compression classify the optimizations into these four classes: 

\begin{itemize}
    \item Compact Model: This technique modifies the standard operations used in DNNs. In a CNN, the standard convolution is replaced by more flexible convolution arithmetics that expand the number of feature maps and decrease the number of parameters such as dilated convolution \cite{yu2016multiscale} or separable depthwise convolution \cite{Chollet_2017_CVPR}.  In an RNN, cells like S-LSTM \cite{s_lstm}, or JANET \cite{janet} simplify the gates and decrease the number of parameters compared to a regular LSTM. 
    
    \item Tensor Decomposition: a tensor is the fundamental data structure used in machine learning. It can represent vectors, matrices and even n-dimensional arrays. Therefore, shrinking the tensors allows accelerating DNNs and reducing their size. Tensor decomposition is an extension of the matrix decomposition techniques used in mathematical settings. Equation \ref{eq:decompose} formulates the matrix decomposition system where the number of parameters of A and B combined is smaller than the number of parameters of M. 
    \begin{equation}
        M = AB\;\text{with}\;M \in \mathbb{R}^{m\times n}, A \in \mathbb{R}^{m\times r}, B \in \mathbb{R}^{r\times n} 
        \label{eq:decompose}
    \end{equation} 
    Note that tensor networks including hierarchical tensor representation (HT) \cite{HT} and tensor train decomposition (TT) \cite{TT} achieve higher compression rates in a fully-connected network since they usually contain more redundancy. 
    
    \item Quantization: In deep learning, quantization~\cite{pmlr-v37-gupta15} refers to converting data objects from 32-floating point to lower precision or a fixed point integer or even binary. These data objects can be the weights of the layer, the activations (the input data's internal representation), the error value, the gradient values and the weight update. Each method differs with the chosen number of bitwidth and the data objects that are quantized.
    
    \item Network Sparsification or pruning~\cite{han2015compression, journals/corr/abs-1710-01878} attempts to compress the model by pruning some weights (edges) or operations (nodes). Usually, the decision of pruning is taken based on its importance, which is directly the weight values or learned via an attention layer.  
\end{itemize}

\subsubsection{HW-NAS}\label{section:back_hw_nas}
Another efficient deep learning technique is HW-NAS. In HW-NAS, we search for the architecture that maximizes the accuracy and hardware usability among a set of architectures. Note that some HW-NAS can be considered under the model compression techniques as they search for the best bitwidth or the best way to prune. We further detail the search for hyperparameters, NN Block or full architectures in Section 4. 

\subsubsection{Code Transformations}\label{section:code_transformation}
An alternative approach that is gaining more attraction these recent years is to apply some code transformations that optimize the DNNs on the operator level~\cite{9222299}. These transformations are hardware-specific and require a compiler to apply the right transformation for the right hardware platform automatically. 

\subsection{Search Algorithms}
In this section, we give a high-level explanation of reinforcement learning and evolutionary algorithms; two of the most used search algorithms in NAS. The explanations are concise to ensure the core ideas are accessible to a broad audience, and so that the entire survey can be read end-to-end easily.  

\subsubsection{Reinforcement Learning} \label{section:back_rl}
Reinforcement learning (RL) is a technique to allow an agent to learn by trial and error. The agent takes actions and interacts with an environment to maximize the accumulative reward. It is usually modelled as a Markov Decision Process (MDP). 

\begin{quote}
    \textit{"Reinforcement Learning is learning what to do - how to map situations to actions -to maximize a numerical reward signal. The learner is not told which actions to take, but instead must discover which actions to yield the most reward by trying them." }\cite{sutton2018reinforcement}
\end{quote}

\begin{figure}[!ht]
    \centering
    \includegraphics[width=7cm]{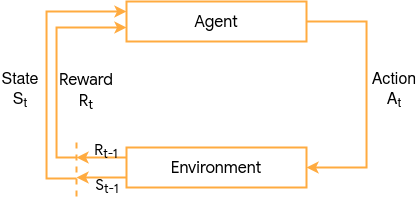}
    \caption{The agent-environment interaction in a Markov Decision Process. Source: \cite{sutton2018reinforcement} }
    \label{fig:rl}
\end{figure}

As illustrated in figure \ref{fig:rl}, at each step, the agent observes the state of the environment sends and receives a reward for its previous action. It then selects its next action. The reward guides the agent to improve its policy such that better actions are chosen in the future. The policy of an agent is the algorithm that allows it to choose between multiple actions.

There are several variants and algorithms to train and update the policy of an agent. The basic one is the \textit{sample-eval-update} method. This method is an iterative process where the agent samples one action, evaluates it using a Q-Learning \cite{q-learning} policy and updates the decision process accordingly. Another update method is \textit{REINFORCE} \cite{reinforce}. It is a Monte-Carlo variant of policy gradients. The agent collects a set of actions (called trajectory) $\tau$ using its current policy, and uses it to update the policy parameter. Since one full trajectory must be completed to construct a sample space, REINFORCE is updated in an off-policy way.

\subsubsection{Evolutionary Algorithms} \label{section:back_ea}
Evolutionary algorithms (EA) are optimization techniques that have three main characteristics: 
\begin{itemize}
    \item Population-based: EA maintain an entire set of candidate solutions. Each solution corresponds to a unique point in the search space of the problem. This set of solutions is called a population. 
    \item Fitness-oriented: EA assign a fitness score to each solution, reflecting the quality of a solution. 
    \item Generations: EA generate new populations by applying a series of transformations to the current population. These transformations are called mutations or crossover operations. Mutations transform one solution into another one, while crossover operations merge two solutions into a new one. 
\end{itemize}

\begin{figure}[!ht]
    \centering
    \includegraphics[width=7cm]{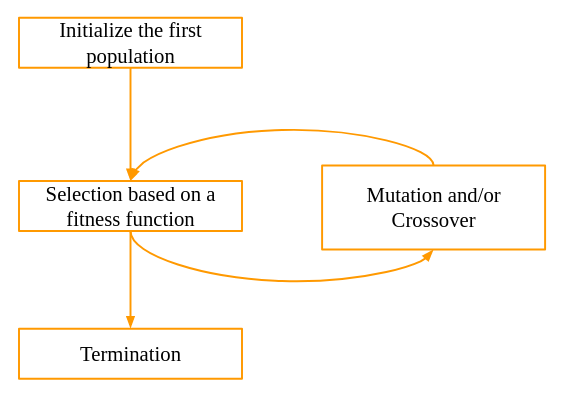}
    \caption{Overview of the evolutionary algorithm steps.}
    \label{fig:ea}
\end{figure}

A genetic algorithm is a type of evolutionary algorithms that encodes the individuals into numerical vectors, called chromosomes. The chromosome is represented as a set of parameters that defines a particular individual.  A selection criterium is used to select a set of candidate individuals of which the fittest are mutated and recombined by crossover to create the next generation. This algorithm is the simplest version of the evolutionary methods as the chromosomes are numerical vectors and the mutations can be simple permutations. 

\section{Taxonomy of HW-NAS}
\label{section:taxonomy}

\begin{figure*}[!t]
    \includegraphics[width=\textwidth]{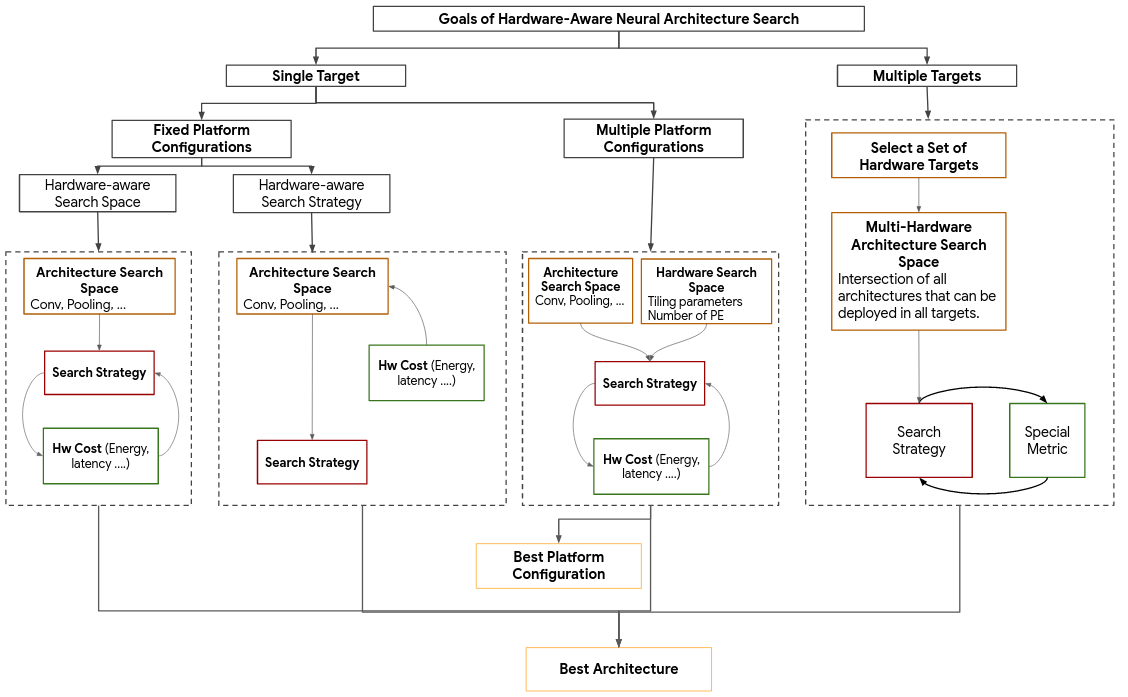}
    \caption{Overview of different hardware-aware NAS designs.}
    \label{fig:goals}
\end{figure*}

\begin{table}[!ht]
	    \centering
	    \begin{tabular}{p{7cm}}
	        \hline
	         Category \& References  \\
	         \hline
	         \textbf{Single Target, Fixed Configuration} \\
	          \hspace*{0.5cm} Hardware-aware Search Strategy \\ 
	          \hspace*{0.5cm} \cite{SmithsonYGM16,morphnet, dong2018ppp-net:, elsken2019efficient, architect, hsu2018monas, tan2019mnasnet, gradientbased, fastneural, nsganet, InstaNAS, tea-dnn, cai2019proxylessnas, wu2019fbnet,chamnet,  sood2019neunets, partial, guo2019single, auto-slim, stamoulis2019single, resourceconstraintxiong, BALDEONCALISTO2020325, han2019design, sparse, tan2019efficientnet, shaw2019squeezenas, Liu_2020_CVPR, nat, bae2019resource,error-resilient, energy-aware, nnoptimize, particle-swarm, zhang2020fast, deepmaker, xu2020latencyaware, iqbal2020flexibo, bian2020nass, chen2020multiobjective, FPNet, edgeinference, fernandezmarques2020searching, memorytext, gupta2020acceleratoraware, huang2020ponas, li2020neural, xiong2020mobiledets, auto-fas, xia2020hnas,phan2020binarizing, CASSIMON2020100234, lee2020neuralscale, wang2020apq, nasmulticognitive, piergiovanni2019tiny, lee2020journey, ferianc2020vinnas, lu2020nsganetv2, hu2020tfnas, 9153122, tsai2020findingtransformers, marchisio2020nascaps, bruggemann2020automated, li2020lcnas, lee2020s3nas, chen2020binarized, niu2020realtime, wang2020efficientnetelite, yuan2020enas4d, apnas, 9201169, lopezdnas, banbury2020micronets, liberis2020munas, yang2020resourceaware}\\
	         \hspace*{0.5cm} Hardware-aware Search Space \\
	          \hspace*{0.5cm}\cite{zhang2020fast, gupta2020acceleratoraware, shufflenasnet,cai2020onceforall, automated} \\
	         \textbf{Single Target, Multiple Configurations} \\
	         \cite{FNAS, fnas2, co-exploration2019, co-exploration20192, abdelfattah2020best, chen2020search, jiang2020standing, lin2020mcunet, colangelo2020automl, choi2020dance, zhang2020dna} \\ 
	         \textbf{Multiple Targets} \\
	         \cite{chu2020discovering, transformable} \\ 
	         \hline
	    \end{tabular}
	    \caption{Taxonomy of Hardware-aware Neural Architecture Search Goals}
	    \label{tab:taxonomy}
	\end{table}

	Unlike conventional NAS, where the goal is to find the best architecture that maximizes model accuracy, hardware-aware NAS (HW-NAS) has multiple goals and multiple views of the problem. We can classify these goals into three categories (See figure \ref{fig:goals} from left to right) : 
	
	\begin{itemize}
		\item \textbf{Single Target, Fixed Configuration}: Most of existing HW-NAS fall under this category. The goal is to find the best architecture in terms of accuracy and hardware efficiency for one single target hardware. Consequently, if a new hardware platform has to be used for the NAS, we need to rerun the whole process and feed it the right values to calculate the new hardware's cost. These methods generally define the problem as a constrained or multi-objective optimization problem \cite{cai2019proxylessnas, tan2019mnasnet, wu2019fbnet}. Within this category, two approaches are adopted: 
		\begin{itemize}
		    \item \textit{Hardware-aware search strategy} where the search is defined as a multi-objective optimization problem. While searching for the best architecture, the search algorithm calls the traditional evaluator component to get the accuracy of the generated architecture but also a special evaluator that measures the hardware cost metric (e.g., latency, memory usage, energy consumption). Both model accuracy and hardware cost guide the search and enable the NAS to find the most efficient architecture. 
		    \item On the other hand, the \textit{Hardware-aware Search Space} approach uses a restricted pool of architectures. Before the search, we either measure the operators' performance on the target platform or we define a set of rules that will refine the search space; eliminate all the architectures' operators that do not perform well on the target hardware. For example, HURRICANE \cite{zhang2020fast} uses different operator choices for three types of mobile processors: Hexagon DSP, ARM CPU and Myriad Vision Processing Unit (VPU). Accumulated domain knowledge from prior experimentation on a given hardware platform help narrow down the search space. For instance, they do not to use depthwise convolutions for CPU, squeeze and excitation mechanisms for VPU and they do not lower the kernel sizes for a DSP. Such gathered empirical information helps to define three different search spaces according to the targeted hardware platform. Note that after defining the search space with these constraints, the search strategy is similar to the one used by conventional NAS, which means that the search is solely based on the accuracy of the architecture and no other hardware metric is incorporated. 
		\end{itemize}
		
		\item \textbf{Single Target, Multiple Configurations}: the goal of this category is not only to get the most optimal architecture that gets the best accuracy but also to get an optimal architecture with latency guranteed to meet the target hardware specification. For example, the authors of FNAS \cite{FNAS} define a new hardware search space containing the different FPGA specifications (e.g., tiling configurations). They also use a performance abstraction model to measure the latency of the searched neural architectures without doing any training. This allows them to quickly prune architectures that do not meet the target hardware specifications. In \cite{NASICS}, the authors use the same approach for ASICs and define a hardware search space that contains various ASIC templates.
		
		\item \textbf{Multiple Targets}: In this third category, the goal is to find the best architecture when given a set of hardware platforms to optimize for. In other words, we try to find a single model that performs relatively well across different hardware platforms. This approach is the most favourable choice, especially in mobile development as it provides more portability. This problem was tackled by \cite{chu2020discovering, transformable} by defining a multi-hardware search space. The search space contains the intersection of all the architectures that can be deployed in the different targets. Note that, targeting multiple hardware specifications at once is harder as the best model for a GPU, can be very different to the best model for a CPU (i.e., for GPUs wider models are more appropriate while for CPUs deeper models are).

	\end{itemize}

\section{Search Spaces}	
{\label{Section:Search_Space}}
Two different search spaces have been adopted in the literature to define the search strategies used in HW-NAS: the {\it{Architecture Search Space}} and the {\it{Hardware Search Space}}.

\subsection{Architecture Search Space}\label{section:arch_ss}
The Architecture Search Space is a set of feasible architectures from which we want to find an architecture with high performance. Generally, it defines a set of basic network operators and how these operators can be connected to construct the computation graph of the model. We distinguish two approaches to design an architecture search space:
\begin{enumerate}
    \item 
 \textit{Hyperparameter Optimization for a fixed architecture:} In this approach a neural architecture is given including its operator choices. The objective is limited to optimizing the architecture hyperparameters (e.g., number of channels, stride, kernel size).
 \item 
 \textit{True Architecture Search Space:} The search space allows the optimizer to choose connections between operations and to change the type of operation.
\end{enumerate}

\begin{figure*}[!ht]
    \centering
    \includegraphics[width=12cm]{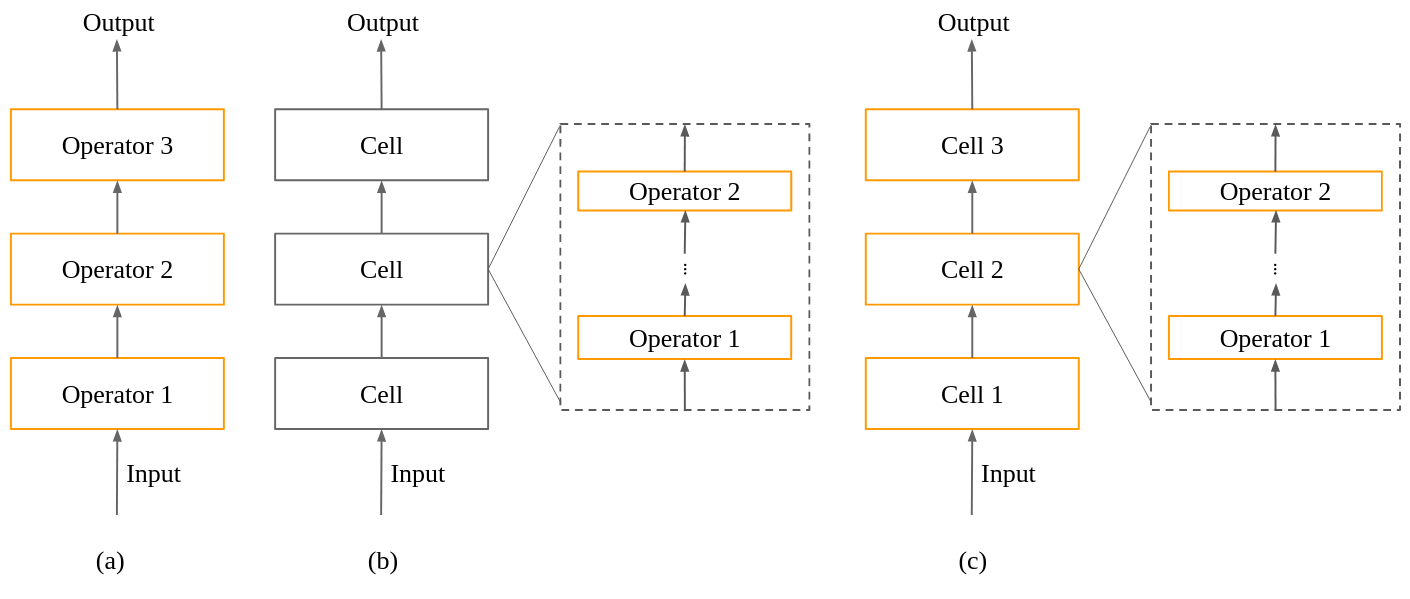}
    \caption{Architecture search spaces types. (a) Global search space, (b) Cell-based search space, and (c) Hierarchical search space. In orange the operators considered during the search.}
    \label{fig:search_space}
\end{figure*}

Both approaches have their advantages and disadvantages but it is worth mentioning that although former approach reduces the search space size, it requires considerable human expertise to design the search space and introduces a strong bias. Whereas the latter approach decreases the human bias but considerably increases the search space size and hence the search time.

Generally, in the latter approach, we distinguish three types (See figure \ref{fig:search_space}): 
\begin{itemize}
	\item \textbf{Layer-wise Seach Space}, where the whole model is generated from a pool of operators. FBNet Search Space \cite{wu2019fbnet}, for example, consists of a layer-wise search space with a fixed macro architecture which determines the number of layers and dimensions of each layer where the first and last three layers have fixed operators. The remaining layers need to be optimized. 
	
	\item \textbf{Cell-based Search Space}, where the model is constructed from repeating fixed architecture patterns called blocks or cells. A cell is often a small acyclic graph that represents some feature transformation. The cell-based approach relies on the observation that many effective handcrafted architectures are designed by repeating a set of cells. These structures are typically stacked and repeated a number of time to form larger and deeper architectures. This search space focuses on discovering the architecture of specific cells that can be combined to assemble the entire neural network.  Although cell-based search spaces are intuitively efficient to look for the best model in terms of accuracy, they lack flexibility when it comes to hardware specialization \cite{tan2019mnasnet, wu2019fbnet}.
	
	\item \textbf{Hierarchical Search Space}, works in 3 steps: First the cells are defined and then bigger blocks containing a defined number of cells are constructed. Finally the whole model is designed using the generated cells. MNASNet \cite{tan2019mnasnet} is a good example of this category of search spaces. The authors define a factorized hierarchical search space that allows more flexibility compared to a cell-based search space. This allows them to reduce the size of the total search space compared to the global search space. 
	
\end{itemize}
In existing NAS research works, the authors define a macro-architecture that generally determines the type of networks considered in the search space. When considering CNNs, the macro architecture is usually identical to the one shown in figure \ref{fig:general_nn}. Therefore, many works \cite{cai2019proxylessnas, wu2019fbnet, tan2019mnasnet, chu2020discovering, liu2019auto} differ in the number of layers, the set of operations and the possible hyperparameters values. Recently, the scope of network type is changing. For instance, NASCaps \cite{marchisio2020nascaps} changes their macro-architecture to allow the definition of capsules. Capsules network \cite{capsules} are basically cell-based CNNs where each cell (or capsule) can contain a different CNN architecture. 

Other works like \cite{nat, tsai2020findingtransformers} focus on transformers and define their macro-architecture as a transformer model. The search consists of finding the number of attention heads and their internal operations. When dealing with the hyperparameters only, the macro architecture can define a variety of network types. Authors in \cite{iqbal2020flexibo, niu2020realtime} mix different definitions, transformers + CNN and transformers + RNN respectively.  They define a set of hyperparameters that encompasses the pre-defined parameters for different network types at the same time. 
	
Lately, more work \cite{cai2019proxylessnas, shaw2019squeezenas} have been considering the use of over-parameterized networks (i.e. supernetworks) to speedup the NAS algorithms. These networks consist of adding architectural learnable weights that select the appropriate operator at the right place. Note that these techniques have been applied to transformers as well \cite{wang2020hat}. 

Finally, in some research efforts, the pool of operators/architectures is refined with only the models that are efficient in the targeted hardware \cite{transformable,chu2020discovering}. The search space size is considerably reduced by omitting all the architectures that cannot be deployed.  

\subsection{Hardware Search Space (HSS)}\label{section:hss}
Some HW-NAS methods include a HSS component which generates different hardware specifications and optimizations by applying different algorithmic transformations to fit the hardware design. This operation is done before evaluating the model. Although the co-exploration is effective, it increases the search space time complexity significantly. If we take FPGAs as an example, their design space may include: IP instance categories, IP reuse strategies, quantization schemes, parallel factors, data transfer behaviours, tiling parameters, and buffer sizes. It is arguably impossible to consider all these options as part of the search space due to the added search computation cost. Therefore, many existing strategies  limit themselves to only few options. 

Hardware Search Space (HSS) can be further categorized as follows:
\begin{itemize}
    \item \textbf{Parameter-based:} The search space is formalized by a set of different parameter configurations. Given a specific data set, \textbf{FNAS} \cite{FNAS} finds the best performing model, along with the optimization parameters needed for it to be deployed in a typical FPGA chip for deep learning. Their HSS consists of four tiling parameters for the convolutions. \textbf{FNASs} \cite{fnas2} extends \textbf{FNAS} by adding more optimization parameters such as loop unrolling. The authors in \cite{co-exploration2019, co-exploration20192} used a multi-FPGA hardware search space. The search consists of dividing the architecture into pipeline stages that can be assigned to an FPGA according to its memory and DSP slices, in addition to applying an optimizer that adjusts the tiling parameters. 
    Another example is presented in \cite{abdelfattah2020best}, where the adopted approach takes the global structure of an FPGA and adds all possible parameters to its hardware search space including the input buffer depth, memory interface width, filter size and ratio of the convolution engine.  \cite{chen2020search} searches the internal configuration of an FPGA by generating simultaneously the architecture hyperparameters, the number of processing elements, and the size of the buffer.
    \textbf{FPGA/DNN} \cite{hao2019fpgadnn} proposes two components: \textit{Auto-DNN} which performs hardware-aware DNN model search and \textit{Auto-HLS} which generates a synthesizable C code of the FPGA accelerator for explored DNNs. Additional code optimizations such as buffer re-allocation and loop fusion on the resulting C-code are added to automate the hardware selection. 
    
    \item \textbf{Template-based:} In this scenario, the search space is defined as a set of pre-configured templates. For example, \textbf{NASAIC}~\cite{NASICS} integrates NAS with Application-Specific Integrated Circuits (ASIC). Their hardware search space includes templates of several existing successful designs. The goal is to find the best model with the different possible parallelizations among all templates. In addition to the the tiling parameters and bandwidth allocation,  the authors in \cite{jiang2020standing} define a set of FPGA platforms and the search finds a coupling of the architecture and FPGA platform that fits a set of pre-defined constraints (e.g., max latency 5ms) 
\end{itemize}

In general, we can classify the targeted hardware platforms into 3 classes focusing on their memory and computation capabilities: 
 \begin{itemize}
     \item Server Processors: this type of hardware can be found in cloud data centers, on premise data centers, edge servers, or supercomputers. They provide abundant computational resources and can vary from CPUs, GPUs, FPGAs and ASICs. When available, machine learning researchers focus on accuracy. Many NAS works consider looking for the best architecture in these devices without considering the hardware-constraints. Nevertheless, many HW-NAS works target server processors to speed up the training process and decrease the extensive resources needed to train a DL architecture and use it for inference.  
     
     \item Mobile Devices: With the rise of mobile devices, the focus has shifted to enable fast and efficient deep learning on smartphones. As these devices are heavily constrained with respect to their memory and computational capabilities, the objective of ML researchers shift to assessing the trade-off between accuracy and efficiency. Many HW-NAS algorithms target smartphones including FBNet \cite{wu2019fbnet} and ProxylessNAS \cite{cai2019proxylessnas} (refer to table \ref{tab:hw_class}). Additionally, because smartphones usually contain system on chips with different types of processors, some research efforts \cite{edgeinference} have started to explore ways to take advantage of these heterogeneous systems.
     
     \item Tiny Devices: The strong growth in use of microcontrollers and IoT applications gave rise to TinyML~\cite{TinyML}. TinyML refers to all machine learning algorithms dedicated to tiny devices, i.e, capable of on-device inference at extremely low power. One relevant HW-NAS method that targets tiny devices is MCUNet \cite{lin2020mcunet}, which includes an efficient neural architecture search called TinyNAS. TinyNAS optimizes the search space and handles a variety of different constraints (e.g., device, latency, energy, memory) under low search costs. Thanks to the efficient search, MCUNet is the first to achieves $>$70\% ImageNet top-1 accuracy on an off-the-shelf commercial microcontroller.
     
 \end{itemize}

 \begin{table}[!ht]
     \centering
     \begin{tabular}{c|p{6cm}}
          Targeted HW & References  \\
          \hline \hline
          CPU & \cite{chamnet, zhang2020fast, xu2020latencyaware, bian2020nass,piergiovanni2019tiny, abdelfattah2020best, edgeinference, fernandezmarques2020searching, 9127125, deepmaker} \\
          \hline 
          GPU & \cite{dong2018ppp-net:, hsu2018monas, InstaNAS, tea-dnn, gradientbased, fastneural, nsganet, sood2019neunets, partial, stamoulis2019single, auto-slim, memorytext, deepmaker, energy-aware, xu2020latencyaware, iqbal2020flexibo, huang2020ponas, piergiovanni2019tiny, lee2020neuralscale, Kim2017NEMON, chen2020search, tabu, BALDEONCALISTO2020325, han2019design, bae2019resource, error-resilient, nnoptimize, particle-swarm,ferianc2020vinnas, lu2020nsganetv2, hu2020tfnas, li2020lcnas, tsai2020findingtransformers, yuan2020enas4d, apnas, 9201169, yang2020resourceaware}\\
          \hline 
          MCU & \cite{lin2020mcunet, sparse, banbury2020micronets, weng2019unet, wu2019fbnet, CASSIMON2020100234}\\
          \hline 
          Mobile Devices & \cite{tan2019mnasnet, cai2019proxylessnas, wu2019fbnet, guo2019single, stamoulis2019single, nat, cai2020onceforall, resourceconstraintxiong, xia2020hnas, xiong2020mobiledets, auto-fas, lopezdnas, li2020neural, fernandezmarques2020searching, tsai2020findingtransformers} \\
          \hline 
          ASICs & \cite{zhang2020fast, co-exploration2019, co-exploration20192, wang2020apq, lee2020s3nas, wang2020efficientnetelite, zhang2020dna, choi2020dance, marchisio2020nascaps}\\
          \hline 
          Edge TPU & \cite{tsai2020findingtransformers, xiong2020mobiledets, gupta2020acceleratoraware}\\ 
          \hline 
     \end{tabular}
     \caption{Classification of HW-NAS based on their targeted Hardware.}
     \label{tab:hw_class}
 \end{table}

\subsection{Current Hardware-NAS Trends} \label{section:hss-trends} 
\begin{figure}[!ht]
    \centering
    \includegraphics[width=8cm]{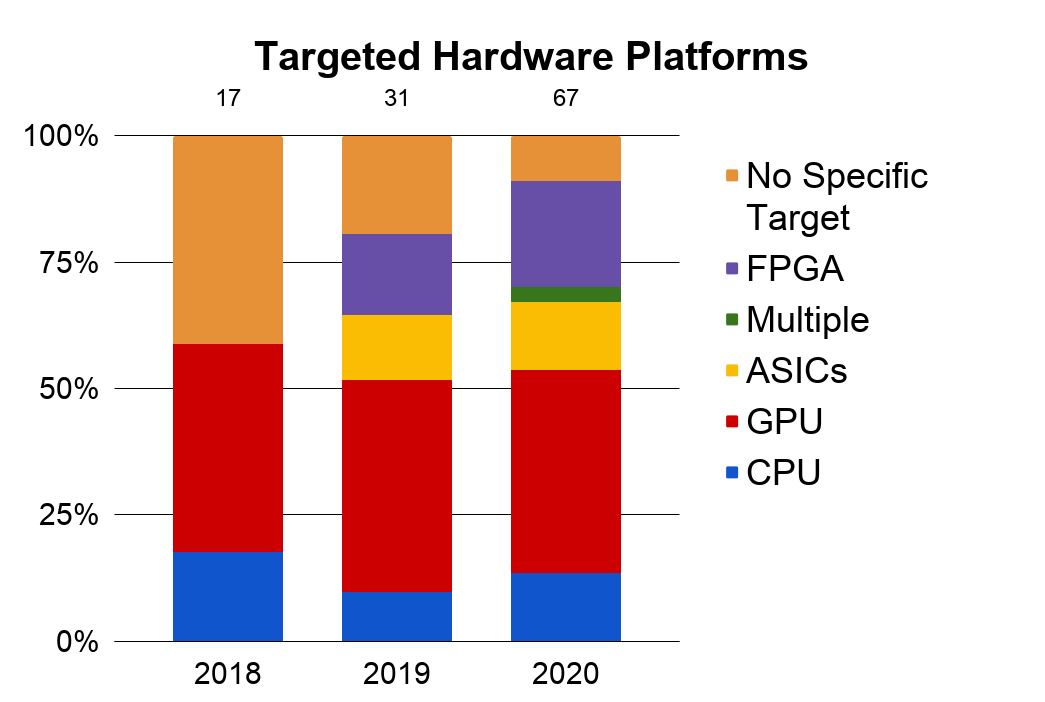}
    \caption{Statistics on targeted platforms}
    \label{fig:hardware}
\end{figure}

Figure \ref{fig:hardware} shows the different types of platforms that have been targeted by HW-NAS in the literature. In total, we have studied 126 original hardware-aware NAS papers. By target, we mean the platform that the architecture is optimized for. Usually the search algorithm is executed in a powerful machine, but that is not the purpose of our study. "\textit{No Specific Target}" means that the HW-NAS incorporates hardware agnostic constraints into the objective function such as the number of parameters or the number of FLOPs. 
 In the figure, the tag "\textit{Multiple}" means multiple types of processing elements have been used in the HW platform. Table \ref{tab:hw_class} gives the list of references per targeted hardware. 
 
\begin{figure}[!ht]
    \centering
    \includegraphics[width=8cm]{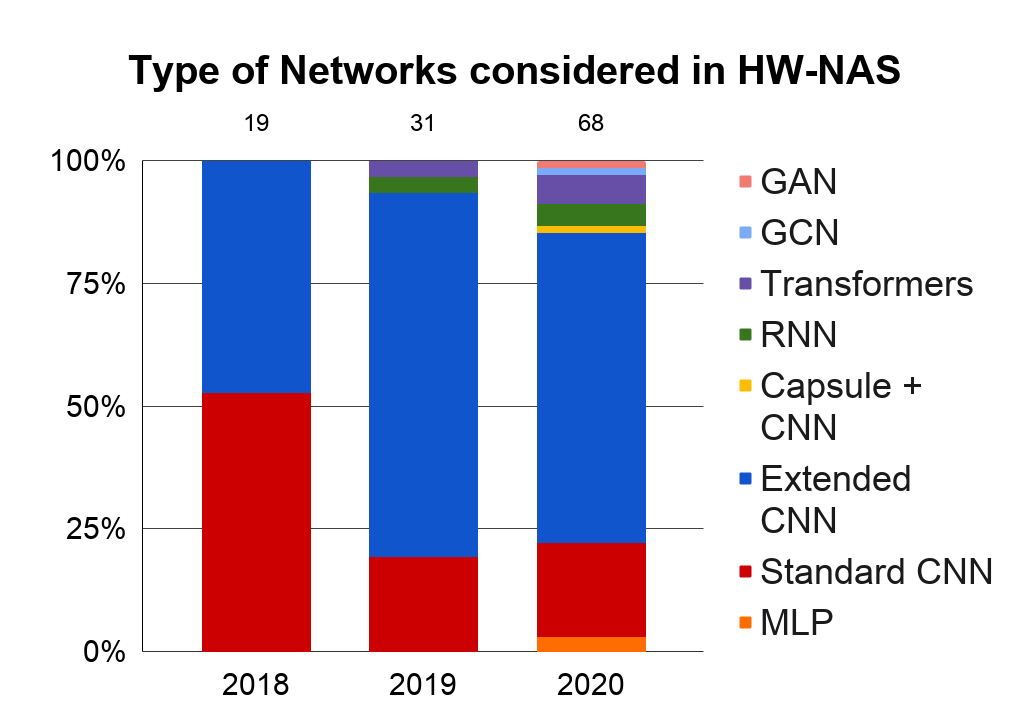}
    \caption{Statistics about the type of networks described by the HW-NAS search spaces}
    \label{fig:networks}
\end{figure}
 
In the figure, we note that the number of research papers targeting GPUs and CPUs has more or less remained constant. However, we can clearly see that FPGAs and ASICs are gaining popularity over the last 3 years. This is consistent with the increasing number of deep learning edge applications. Two recent interesting works are \cite{chu2020discovering, transformable} both of which target multiple hardware platforms at once.

 In figure \ref{fig:networks}, we illustrates the different DNN operations that compose the architecture search space. First, we divide the CNN into two groups, \textit{standard CNN} which only utilizes a standard convolution and \textit{extended CNN} which involves special convolution operations such as the depthwise separable convolution or grouped convolutions. NAS has been mostly dominated by convolutional neural networks as shown in the figure. However, recent works have started explore more operators by incorporating capsule networks \cite{marchisio2020nascaps}, transformers \cite{klyuchnikov2020nasbenchnlp}, and GANs \cite{lee2020journey}.


\section{Hardware-aware NAS Problem Formulation} \label{Section:HW_NAS_Formulation}
Neural Architecture Search (NAS) is the task of finding a well-performing architecture for a given dataset. It is cast as an optimization problem over a set of decisions that define different components of deep neural networks (i.e., layers, hyperparameters). This optimization problem can simply be seen as formulated in equation \ref{eq:NAS}. 

\begin{equation}
	\max_{\alpha\in A} f(\alpha, \delta)
	\label{eq:NAS}
\end{equation}

We denote the space of all feasible architectures as A (also called search space). The optimization method is looking for the architecture $\alpha$ that maximizes the performance metric denoted by $f$ for a given dataset $\delta$. In this context, $f$ can simply be the accuracy of the model. 

Although it is important to find networks that provide high accuracy, these NAS algorithms tend to give complex models that cannot be deployed on many hardware devices. To overcome this problem, practitioners consider other objectives, such as the number of model parameters, the number of floating-point operations, and device-specific statistics like the latency or the energy consumption of the model. Different formulations were used to incorporate the hardware-aware objectives within the optimization problem of neural architecture search. We classify these approaches into two classes, single and multi-objective optimization. The single objective optimization can be further classified as two-stage or constrained optimization. Similarly, the multi-objective optimization approach can be further classified as single or multi-objective optimizations. Please refer to figure~\ref{fig:problem_form} for a summary of these approaches. These 2 classes are further detailed with examples from the literature in the following sections: 

\begin{figure*}[!ht]
    \centering
    \includegraphics[width=14cm]{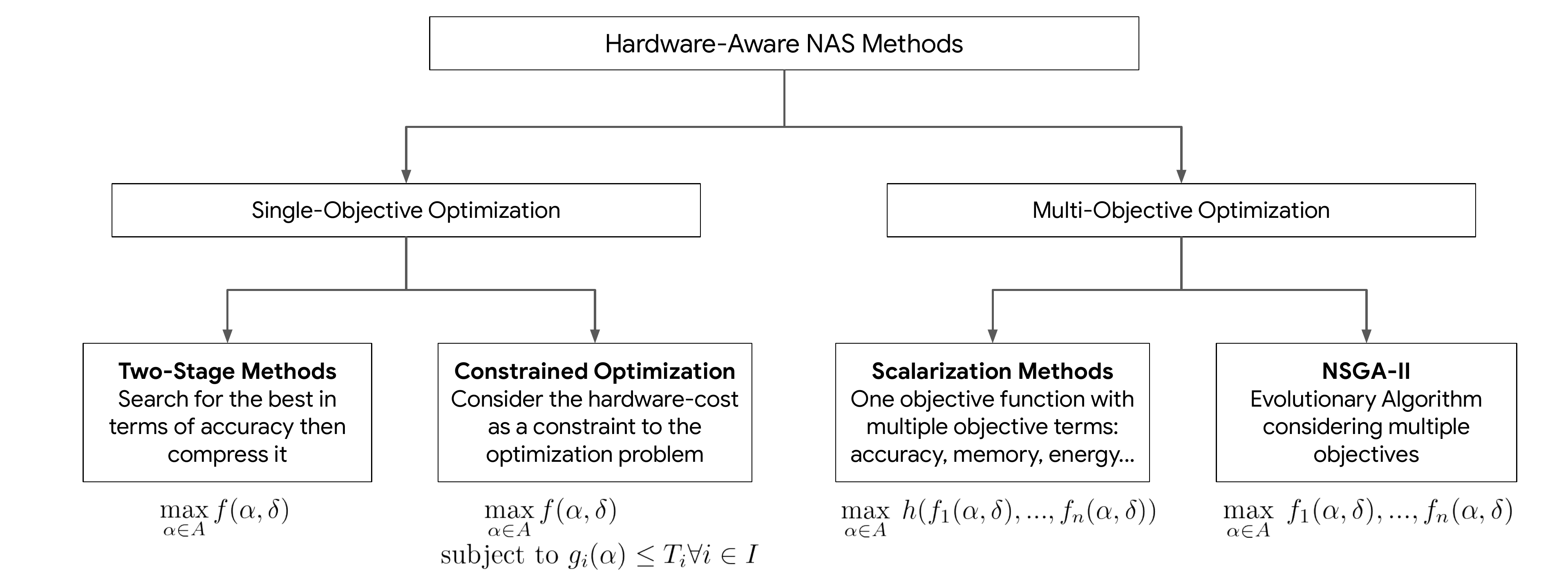}
    \caption{HW-NAS problem formulations.}
    \label{fig:problem_form}
\end{figure*}

\subsection{Single-Objective Optimization}
 In this class, the search is realized considering only one objective to maximize, i.e the accuracy. Most of the existing work in the literature \cite{tan2019mnasnet, cai2019proxylessnas, FPNet, jiang2020standing, abdelfattah2020best}, that tackle the hardware-aware neural architecture search, try to formulate the multi-objective optimization problem into a single objective to better apply strategies like reinforcement learning or gradient-based methods. We can divide this class into two different approaches: Two-stage optimization and constrained optimization. 

\subsubsection{Two-Stage optimization} \label{section:two-stage} In this first category, we retain the original formulation of the NAS problem and then specify the model for deployment. This approach is suboptimal as the final architecture proposed by the NAS is not always the one that gives the best performances on the hardware device. Two-stage optimization consists of applying NAS methods to obtain the best-performing architecture and then in a second stage, specialize this architecture for deployment on a target hardware platform. This specialization performs a series of optimization to fit the hardware requirements. \cite{han2019design} applies a reinforcement learning agent to find the best quantization bitwidth and pruning level after selecting the most accurate model. \cite{cai2020onceforall} searches over a pre-trained and selected architecture to find the most efficient one in terms of latency and energy consumption. 

\subsubsection{Constrained optimization} \label{section:constrained} In this approach, the hardware-aware characterizations are considered as constraints in the original NAS formulation. The constraints take the form of thresholds to be respected. For example, inference time, energy consumption, memory occupation, etc. The conditions are added as constraints to the optimization problem to enforce requirements like fewer parameters or faster inference time. The threshold and the trade-off between different constraints can be adapted to practical requirements. For such cases, the single-objective optimization problem defined in Equation \ref{eq:NAS} turns into a constrained optimization problem defined by 

\begin{equation}
	\begin{split}
		& \max_{\alpha\in A} f(\alpha, \delta) \\
		&\text{subject to}\; g_{i}(\alpha) \leq T_i \; \forall i \in I
	\end{split}
	\label{eq:constrained}
\end{equation}

Here, $g_i$ corresponds to the different constraints taken into account (e.g., latency, memory, energy consumption) and $T_i$ denotes the respective threshold. 
As most of the optimization methods used by NAS (i.e. reinforcement learning and evolutionary algorithms) were designed for unconstrained optimization problems, this formulation is hard to be adopted directly. Therefore, many researchers turned to penalty methods to transform the equation into a single objective function that contains the hardware constraints as well as the accuracy measurement \cite{tan2019mnasnet, wu2019fbnet, cai2019proxylessnas}. 
For example, MNASNet \cite{tan2019mnasnet}, uses equation (\ref{eq:mnasnet}), where $f$ is the accuracy measurement function, LAT is the latency of the model and $T$ is the threshold. They use a learnable parameter $w$ to control the effect of the hardware constraints on the global objective function.

\begin{equation}
	\max_{\alpha\in A}\; f((\alpha) \cdot [LAT(\alpha) / T]^w 
	\label{eq:mnasnet}
\end{equation}

ProxylessNAS \cite{cai2019proxylessnas} uses a loss function that comprises of the cross-entropy (CE) loss and hardware-aware constraints. 

\begin{equation}
	\mathcal{L} = \mathcal{L}_{\text{CE}} + \lambda_1 ||w||^2
	+ \lambda_2 E[\text{latency}]
	\label{eq:proxylessnas}
\end{equation}
Equation \ref{eq:proxylessnas} illustrates the loss calculated by the reinforcement learning agent used by ProxylessNAS. $\lambda_1$ and $\lambda_2$ are learnable parameters that adjust the effect of the efficiency of the overall loss. Specifically, a  policy  is  learned  that decides whether to add, remove or keep a layer as well as whether to alter its number of filters.

\subsection{Multi-Objective Optimization}

Another approach is to handle multiple fronts in the formalism of a multi-objective optimization problem defined as: 

\begin{equation}
	\max_{\alpha\in A}\; f_1(\alpha, \delta), f_2(\alpha, \delta) , ..., f_n(\alpha, \delta)
	\label{eq:multi}
\end{equation}
In this scenario, there is often no single optimal solution that simultaneously maximizes every objective function. Additionally, some objective functions can be conflicting. For instance, trying to minimize the number of parameters while aiming at maximize the accuracy. Therefore, in these situations, the task boils down to finding Pareto-optimal solutions. These techniques can: \begin {enumerate}
	\item transform the problem to a single-objective optimization using scalarization method, also called weighted sum method or, 
	\item solve the multi-objective optimization problem using dedicated heuristics or meta-heuristics such as genetic algorithms or tabu search {\cite{tabu}}. In general, this second approach provides not one optimal solution but a set of solutions that form the optimal Pareto front of the multi-objective optimization problem. 
	\end{enumerate}

\subsubsection{Scalarization Methods}\label{section:scalarization}  One way to solve the multi-objective optimization problem is to use a scalarization approach. Equation \ref{eq:scalarization} formulates this method. We use a parameterized aggregation function $h$ to transform the multi-objective optimization problem into a single-objective optimization problem. 

\begin{equation}
	\max_{\alpha\in A}\; h(f_1(\alpha, \delta), f_2(\alpha, \delta) , ..., f_n(\alpha, \delta))
	\label{eq:scalarization}
\end{equation}

The function $h$ can be a weighted sum, a weighted exponential sum, a weighted min-max or a weighted product. However, in most situations, not all Pareto optimal solutions can be found in solving this problem with a fixed setting of the weights. Therefore, the problem is solved for multiple values of the vector $w$ which requires multiple optimization runs. To mitigate the cost of having multiple run, researchers commonly use a set of fixed weights according to the desired trade-off between the objectives and the practitioners’ preferences.
Thanks to the scalarization, the problem becomes a single-objective optimization problem which can be solved by any optimizing methods discussed in Section \ref{section:SearchStrategy}.
For example, \cite{hsu2018monas} proposes to use the weighted sum as the objective function. The proposed formulation of this function is described by equation \ref{eq:monas}. \textit{ACC} refers to the accuracy metric, \textit{E} refers to the energy consumed by the architecture $\alpha$ and \textit{w} is a learned parameter to adjust the effect of the energy on the reward function. 

\begin{equation}
	\max_{\alpha\in A}\; w\cdot ACC(\alpha, \delta) - (1 - w)\cdot E(\alpha) 
	\label{eq:monas}
\end{equation}

\subsubsection{NSGA-II} 
\label{section:nsga} 
An alternative approach is to use the elitist evolutionary algorithm NSGA-II \cite{deb2002fast} which claims that a linear combination of objectives is suboptimal. 
Thus, it is important to cast the problem as a multi-objective optimization problem, where a series of models is found along the Pareto front of multiple objectives such as accuracy, computational cost or inference time, and number of parameters. 
In this algorithm, the architectures are divided into fronts based on their dominance. The architecture in the $i$-th front is only dominated by all the architectures in the $1,\dots, i-1$ fronts. 
Within each front, the architectures are prioritized by the crowding distance, which is computed by the sum of all the neighborhood distances across all the objectives.  Hardware-aware NAS works \cite{Kim2017NEMON, nsganet,chu2019multi} have been using NSGA-II algorithm to ensure the exploration of diverse architectures in the search space. Moreover, NSGANet \cite{nsganet} uses Bayesian Optimization to profit from search history. MoreMNAS \cite{chu2019multi} uses a hybrid search strategy combining NSGA-II with reinforcement learning to regulate arbitrary mutations. 

\section{Search Strategies}
\label{section:SearchStrategy}
In this section, we will review all the search strategies used in hardware-aware NAS and how the hardware cost is integrated into each one of them. Following the algorithm presented in \ref{alg:1}, the search strategy must define:
\begin{itemize}
    \item An accuracy evaluation method: usually it is the actual training of the architecture on a subset of the dataset provided by the user $D_{train}$. However, training each architecture while searching is computationally expensive and takes several GPU hours. Therefore, some speedup methods that try to decrease the training time and estimate the accuracy without effectively training an architecture are used and presented in section \ref{section:speedup}. 
    \item A hardware cost evaluation method: the hardware metrics need to be measured either by a real-time execution of each architecture on the targeted platform or an estimation method. The approaches used are presented in section \ref{section:hw_cost}.
    \item A search algorithm: The search algorithm defines how the architectures are sampled from the search space, either with a generator or at random, and how the search explores this search space by updating the sampling strategy to better fit the best models according to an objective function. See the next section \ref{section:sa}. 
\end{itemize}

\begin{algorithm}[!ht]
\SetAlgoLined
\KwIn{The Search Space $S$, a Dataset $D$, Accuracy Evaluation Methods $E_{acc}(S, w(S), D)$, Hardware Evaluation Methods $E_{hw}(S, w(s), D)$, an Aggregation function $f$}
\KwResult{Optimal Deep Learning Architecture }
    \Repeat{convergence or time limit is reached}{
      Sample an architecture $s$ from the search space $S$;
      Train $s$ on $D$ to obtain the trained weights $w(s)$;
      Compute $acc = E_{acc}(s, w(S), D)$; 
      Compute $hw\_cost = E_{hw}(s, w(s), D)$;
      
      \uIf{$f(acc, hw_cost)$ surpass the best seen value}{
        Update best architecture $<s, w(s))$; 
      }
      Update sampling algorithm with $f(acc, hw_cost)$;
    }
 \caption{A generalized pipeline of HW-NAS}
 \label{alg:1}
\end{algorithm}

\subsection{Search Algorithm}{\label{section:sa}}
\begin{table}[!t]
    \centering
    \begin{tabular}{p{4cm} p{4cm}}
        \hline
         Strategy & References  \\
         \hline
         \textbf{Evolutionary Algorithm} & \cite{elsken2019efficient, nsganet, chamnet, sood2019neunets, fnas2, error-resilient, BALDEONCALISTO2020325, deepmaker, phan2020binarizing, xia2020hnas, wang2020apq, piergiovanni2019tiny, lu2020nsganetv2, nasmulticognitive, lin2020mcunet, marchisio2020nascaps, colangelo2020automl, wang2020efficientnetelite, 9201169} \\
         
         \textbf{Reinforcement Learning} & \cite{architect, hsu2018monas, tan2019mnasnet, fastneural, InstaNAS, shufflenasnet, FNAS, auto-slim, resourceconstraintxiong, han2019design, co-exploration2019, bae2019resource, nat, nnoptimize, co-exploration20192, bian2020nass, chen2020multiobjective, FPNet, NASICS, abdelfattah2020best, gupta2020acceleratoraware, xiong2020mobiledets, chen2020search, CASSIMON2020100234, lee2020neuralscale, lee2020journey, 9127125, jiang2020standing, chu2020discovering, niu2020realtime, apnas} \\
         
         \textbf{Gradient-based Methods} & \cite{zhang2018search, cai2019proxylessnas, wu2019fbnet, guo2019single, single2, stamoulis2019single, shaw2019squeezenas, cai2020onceforall, fastneural, xu2020latencyaware, memorytext,huang2020ponas, li2020neural, auto-fas, zhang2020fast, transformable, ferianc2020vinnas, hu2020tfnas, tsai2020findingtransformers, bruggemann2020automated, li2020lcnas, lee2020s3nas, chen2020binarized, choi2020dance, yuan2020enas4d, lopezdnas, banbury2020micronets, zhang2020dna} \\
         
         \textbf{Bayesian Optimization} & \cite{tea-dnn, sparse, iqbal2020flexibo, liberis2020munas, 2020arXiv200707743G} \\ 
         
         \textbf{Random Search} & \cite{automated} \\ 
         \multirow{2}{3cm}{\textbf{Hybrid}} & EA + RL  \cite{chu2019multi, fastaccuratelight} \\
         & EA + Bayesian Optimization  \cite{yang2020resourceaware} \\
         & EA + Gradient-based\cite{energy-aware} \\
         \hline
    \end{tabular}
    \caption{Search Strategies of HW-NAS}
    \label{tab:strategies}
\end{table}

\subsubsection{Reinforcement Learning}\label{section:rl} 
Most HW-NAS methods use reinforcement learning to search for the best architecture \cite{tan2019mnasnet, FPGA, NASICS, abdelfattah2020best, jiang2020standing, bender2020weight, co-exploration20192, FPNet} because the NAS problem is easily modeled as a Markov Decision Process. The RL controller samples an architecture from the search space and is rewarded according to its accuracy and hardware cost. The agent will then adjust its weights to generate better models. Different works differ on how they represent the agent's policy (set of actions) and how they optimize it. 

Using reinforcement learning, \textbf{MNASNet} \cite{tan2019mnasnet} tries to find the Pareto optimal solution of the objective function described in equation \ref{eq:mnasnet}. It uses a \textit{sample-eval-update} loop to train its RNN controller. To generate a block, the controller will first choose two hidden states (i.e., outputs of previous blocks) as inputs. Then, it will select an operation to apply to each one of them. Finally, it selects a combination method (e.g., addition or concatenation) to obtain the final output of the block. This implementation has been defined by NASNet \cite{zoph2018learning} before. 
Once a model is sampled, it is trained on the target task to get its accuracy and deployed on real phones to get its latency. The system computes the reward value and adjusts the controller parameters accordingly. 

A similar approach was used by \textbf{FPNet} \cite{FPNet} but the RNN controller predicts only the architectural hyperparameters (i.e., number of filters, filter height, filter width, stride height, stride width. etc.) while keeping The macro architecture fixed. 

\textbf{Codesign-NAS} \cite{abdelfattah2020best} proposes to use reinforcement learning to explore both architecture and hardware search spaces. The authors investigated three RL-based search strategies and use the REINFORCE algorithm to improve the accuracy and efficiency of image classification on FPGA. The first strategy consists of combining the two search spaces and updating the CNN and accelerator options at the same time. The second method uses two different controllers, one to learn the CNN architecture and another one to select the best FPGA options. The last one is a conventional NAS where they separately search for the best model in terms of accuracy then as a completely separate step look for the most efficient model. An expected result is that this latter strategy gives bad results when it comes to the constrained environment. An interesting result is that the second method (i.e., phase search) seems to be the most promising achieving higher rewards in most of their experiments.

\subsubsection{Evolutionary Algorithm} \label{section:ea} 
Another popular strategy in conventional NAS \cite{zoph2017neural, real2017largescale} is the use of evolutionary algorithms. Generally, neuro-evolutionary NAS evolves a population of models, sample some models to generate offsprings by applying some mutations (recombination is not used in neuro-evolutionary NAS), and finally evaluate the fitness of the offsprings and update the new generation by adding the best ones to the population. When it comes to integrating the hardware constraints to the NAS algorithms, some research efforts have used evolutionary algorithms \cite{wang2020hat, lin2020mcunet, marchisio2020nascaps, zhang2020fast}. 

\textbf{TinyNAS} \cite{lin2020mcunet} uses a hardware-aware search space approach. It first optimizes the search space to fit the tiny and diverse resource constraints and then performs an evolution search algorithm within the optimized search space to find the most accurate model. The evolution search is performed on a trained super network that contains all the possible sub-networks. First, they sample 100 satisfying networks that fit the resource constraints. Then, just like conventional NAS, they measure the validation accuracy of each model, mutate the offsprings and update the new generation. This process is repeated for 30 iterations.  

\textbf{Once-for-all} \cite{cai2020onceforall} proposes a special technique to train the supernetwork using gradient-based methods called \textbf{progressive shrinking}. A supernetwork is another name for the over-parameterized network. In their version, the supernetwork includes the highest values for the width, depth and channel number in each convolution. 
The process begins by training for larger width, depth and channel numbers and then fine-tune the network for the smaller sizes. Once the supernetwork is trained, they specialize the network using an evolutionary algorithm for different hardware settings with a combined cost of accuracy and latency.

\textbf{HAT} \cite{wang2020hat} is interesting because to our knowledge it is the only work that is trying to search for efficient transformers, targeting NLP tasks. The authors perform an evolutionary search with hardware latency constraints on a SuperTransformer. This means the engine only adds SubTransformers with latency smaller than the hardware constraint to the population.

\textbf{NASCaps} \cite{marchisio2020nascaps} proposes a NAS framework that generates Capsule Networks along with CNN. The proposed framework uses a multi-objective Genetic Algorithm (based on NSGA-II, see section 4.2) to pick the Pareto optimal solutions. The two key operations are the \textit{crossover and mutation}. In the crossover, they define the splitting point by ensuring that the generated DNN is made up of at least on initial convolution layer and a minimum of two capsules.  No standard convolution is placed between two capsule layers. The mutation is performed by randomly choosing one of the layer descriptors from the candidate network and modifying one of the main parameters of the selected layer. 

\subsubsection{Gradient-Based Methods} \label{section:gradient} 
Arguably the most promising search strategy promising  in  terms  of  results, Gradient-based methods are increasingly used by hardware-aware NAS \cite{cai2019proxylessnas, wu2019fbnet, transformable, shaw2019squeezenas, nayman2019xnas} and NAS generally. Running the search separated from the evaluation requires a lot of time and computation. Therefore, a common idea is to have a supernetwork that can emulate any child model in the search space. This means that different parts of the graph share weights between their common edges. This idea of \textit{weight sharing} has the advantage of considerably reducing the search time. Gradient-based methods train the supernetwork to simultaneously get the architecture parameters and weights. This technique has been initiated by DARTS \cite{liu2019darts}.

\textbf{ProxylessNAS} \cite{cai2019proxylessnas} is one of the papers pioneering this method. It defines a supernetwork with binary architecture parameters (i.e., 1 implies that the operator is selected in the layer and 0 otherwise). The loss function combines the cross-entropy and the latency to better update the weights and architecture parameters. An approach similar to BinaryConnect \cite{courbariaux2016binaryconnect} is used to update the binary architecture parameters using an approximation of the gradient w.r.t architecture parameters. 

\textbf{FBNet} \cite{wu2019fbnet} proposes to use differentiable neural architecture search to discover hardware-aware efficient CNNs. They also use a combination of the cross-entropy and latency to train their supernetwork using stochastic gradient descent. Rather than using binary parameters, they relax the problem of finding the best architecture to finding a distribution that yields to the best model.

\textbf{SqueezeNAS} \cite{shaw2019squeezenas} focuses on semantic segmentation and uses a method very similar to FBNet \cite{wu2019fbnet}. \textbf{HTAS} \cite{transformable} uses a gradient-based method to find the best width and depth for their transformable CNNs. 

\textbf{XNAS} \cite{nayman2019xnas} proposes to use the prediction with expert advice theory \cite{predictionEA} for the selection. It leverages the Exponentiated-Gradient algorithm (EG) \cite{kivinen1997exponentiated} rather than the classical gradient descent which prevents the decay of architecture weights to promote the selection of arbitrary architectures.  
    	
\subsubsection{Random Search \& Bayesian Optimization} \label{section:rs_bo} 
The most convenient and easiest search strategy to implement is the random search strategy. Generally, random search and Bayesian optimization are used for hyperparameter optimization. Therefore, most of the existing works that have adopted random search optimize the architectural hyperparameters within a fixed macro architecture. They argue that it is more important to design the architecture search space for the targeted hardware platform than to complicate the search strategy and incorporate the hardware constraints in the objective function. NASNet \cite{nasnet}, for example, tried both methods (i.e., reinforcement learning and random search). They found that with reinforcement learning the results are slightly better (Top-1 accuracy on CIFAR-10, 0,912 - 0,925). 

Li et al. \cite{li2020random} investigate the use of random search on two standard NAS benchmarks (i.e., PTB and CIFAR-10). They use an approach that is similar to ProxylessNAS  \cite{cai2019proxylessnas} which allows them to train a single network at a time and thus reduce the memory footprint with weight sharing. As a result, they show that random search with early-stopping is a competitive NAS baseline as it outperforms ENAS \cite{enas}. Furthermore, random search with weight-sharing outperforms random search with early stopping, achieving SOTA results on PTB. A more recent investigation \cite{stamoulis2019single} showed that random search cost time is not negligible and comparable to NAS methods. 

\subsection{Non-differentiable Hardware Constraints} \label{section:non_diff} 
In many of these search strategies, the authors use an over-parameterized network. Therefore, the loss function must be differentiable w.r.t the architectural parameters. Also, we need to check that the incorporated hardware cost is differentiable. Several methods have been used to make the gradient computation over discrete variables possible. In this section, we review some extensions to the differentiable architecture search space \cite{liu2019darts}. 

\paragraph{Gumbel Softmax} \cite{jang2016categorical} One way to relax the discrete variables is to use the Gumbel softmax function. It helps insert some random noise following the Gumbel distribution so that the gradient computation is possible. This technique was used by FBNet \cite{wu2019fbnet, wu2018mixed}. 

\paragraph{Estimated Continuous Function} ProxylessNAS \cite{cai2019proxylessnas} mimics the concept of BinaryConnect \cite{courbariaux2016binaryconnect}. They approximately estimate the gradient w.r.t the architecture parameters using the gradient w.r.t the binary gates. To reduce the computational cost, they compare the gates two-by-two by factorizing the task of using one out of N paths into multiple binary selection tasks. 

\paragraph{REINFORCE algorithm}
An alternative approach to BinaryConnect is also proposed by ProxylessNAS \cite{cai2019proxylessnas}. They utilized REINFORCE to train the binarized weights. Furthermore, they combine the gradient-based update rule to the REINFORCE updates to form a new general update rule for the architecture parameters. 

\subsection{Runtime Performance Optimization Strategies}
\label{section:speedup}
In this section, we discuss the various methods that are used to speed up the NAS algorithms. We are not going to mention weight sharing as it was described in section \ref{section:gradient}, but we note that it is another way to accelerate the process tied to the search strategy used. 

While NAS methods are efficient at finding state of the art architectures, their search cost is extremely high. Undoubtedly, the most time-consuming component of the NAS process is the training of each model to obtain the validation accuracy and thus making an evaluation. This training alone requires hours for just one architecture on a GPU. For example, training ResNet-50 requires 29 hours on 8 Tesla P100 GPUs \cite{tanakaimagenet}. For NAS, we want to estimate which architecture(s) is the most accurate without needing the exact accuracy. Therefore, we will review the accuracy estimation methods.   

\paragraph{Early Stopping}\label{section:early_stopping} 
Early stopping is a technique that is used by several NAS works \cite{tan2019mnasnet, cai2019proxylessnas, li2020random}. The idea is to train the models for few epochs (usually five epochs) and to take the validation accuracy of this premature model as an approximation of the performance of the fully trained model. 

\paragraph{Hot start} \label{section:hot_start} 
HotNAS \cite{jiang2020standing} proposes the idea of hot starting. The search heuristics does not start from a random model but it starts with an efficient model. Starting from ProxylessNAS \cite{cai2019proxylessnas} and MNASNet \cite{tan2019mnasnet} models, they obtained models with better latency and slightly lower accuracy: 91.47 and 92.2 for HotNAS and ProxylessNAS repectively. In addition to obtaining efficient models, the NAS search is faster as we start from a suboptimal model.

\paragraph{Proxy datasets} \label{section:proxy} 
Using proxy datasets like CIFAR-10 helps train the model with fewer data elements and then fine-tune it to be used for higher resolutions and more complex tasks. 

\paragraph{Accuracy Prediction Models} \label{section:accuracy_prediction} 
Another approach to speed up the training process is to directly predict the accuracy based on the architecture specifications and dataset characteristics. The most used prediction models are the following. 

\textbf{Peephole} \cite{deng2017peephole} is a framework to predict network performance before training based on its architecture. It encodes the architectures into vectors and inputs them onto an LSTM layer. The result is concatenated to the number of epochs and passed to an MLP which outputs the predicted accuracy. Similarly, PNAS \cite{liu2018progressive} uses an RNN model to handle variable-sized inputs which consist of multiple LSTM layers. The final LSTM hidden state goes through a fully-connected layer and sigmoid layer to regress the validation accuracy. 

Another approach by \cite{baker2017accelerating} uses a simple regressio model (SRM regressor) with features like the number of weights and the number of layers as well as the hyperparameters,  to predict the learning curves during the training process.

\textbf{NeuNetS} \cite{sood2019neunets} defines a special component called \textit{TAP}, which stands for Train-less Accuracy Predictor. TAP is designed to perform fast and reliable  CNN  accuracy  predictions. 

Recently, \cite{tang2020semi} uses a semi-supervised performance predictor. To locate latent feature representations, the architecture graph representation is used by an auto-encoder, which is then fine-tuned using a graph similarity calculation. Finally, a graph convolution neural network is used to output the final performance. 


\section{Hardware Cost Estimation Models}\label{section:hw_cost} 
An important component in HW-NAS is the hardware cost measurements. First of all, many metrics have been used in order to evaluate the hardware efficiency of an architecture including the number of FLOPs, the number of parameters, the latency or execution time of the inference, the energy consumption, the memory footprint, the area of the hardware platform, etc.

\paragraph{FLOPs \& Model Size} The first HW-NAS approaches that were published in 2016 and 2017 \cite{SmithsonYGM16, morphnet} use the number of paramaters and number of FLOPs as an objective function to minimize. These techniques assume that the number of operations is positively correlated to the execution time. However, recent works have proved that two models can have the same number of FLOPs but different latencies \cite{bouzidi2020performance, zhang2020fast, wang2020hat}. For example, NASNet-A and MobileNetV1 have roughly similar number of FLOPs, yet, NASNet-A can have slower latency due to the hardware-unfriendly structure. Therefore, using FLOPs as the hardware cost metric is not efficient and may return suboptimal models. On the other hand, using the model size represented by the number of parameters allows to reduce the memory footprint and tends to be considered as an automatic method to search for compressed models \cite{wang2020apq}. 

\paragraph{Latency} Searching for low-latency architectures at inference time is crucial for real world application such as autonomous driving and traffic control. Moreover, resource-limited devices have latency constraints. Thus, a lot of works consider the latency in their objective function and search for the trade-off between inference time and accuracy. 

\paragraph{Energy Consumption} Energy is usually profiled by the provided hardware platform profilers such as nvprof by NVIDIA. The energy can be formalized either as the peak power consumption or the average power, both metrics are used by different HW-NAS works including \cite{hsu2018monas, netadapt, gong2019mixed}. 

\paragraph{Area} Another metric that interest chip manufacturers is its area. The goal is to get the smallest chip possible that could run the best model. \cite{NASICS} uses MAESTRO \cite{maestro} to explore the area and power consumption and search for the best model and best ASICs templates within a set of pre-defined ones. The area of the circuit is also a good indicator of the static power consumption. These two values are correlated.

\paragraph{Memory footprint} We can get the model size by calculating the number of parameters it needs to learn but a more efficient way is to profile how much memory it uses while running; this is the memory footprint. Having a low memory footprint is important for edge devices. These devices are not able to run models with high memory footprint. To reduce the memory footprint in edges devices, techniques like those presented in Section {\ref{Section:Other_Considerations}} are applied.  

\begin{table}[!t]
    \centering
    \begin{tabular}{|p{1cm}|p{1cm}|p{1cm}|p{1cm}|p{1cm}|p{1cm}|}
        \hline
          & Real-time measurements & Lookup Table & MLP & XGBoost & Analytical Estimation \\ \hline
         RMSE & 0 & 4.32 & 3.83 & 3.8 & 5.6 \\ \hline 
         Search Time (s) & 172200 & 74869 & 31713 & 30531 & 33123 \\ \hline
         Search Time Speedups & 1 & 2.3 & 5.43 & 5.64 &  5.2 \\ \hline
    \end{tabular}
    \caption{Comparison of Hardware Cost Measurement Methods. The hardware cost is measured on Tesla K80 GPU.}
    \label{tab:hw_cost_comp}
\end{table}

\begin{figure}[!ht]
    \centering
    \includegraphics[width=9cm]{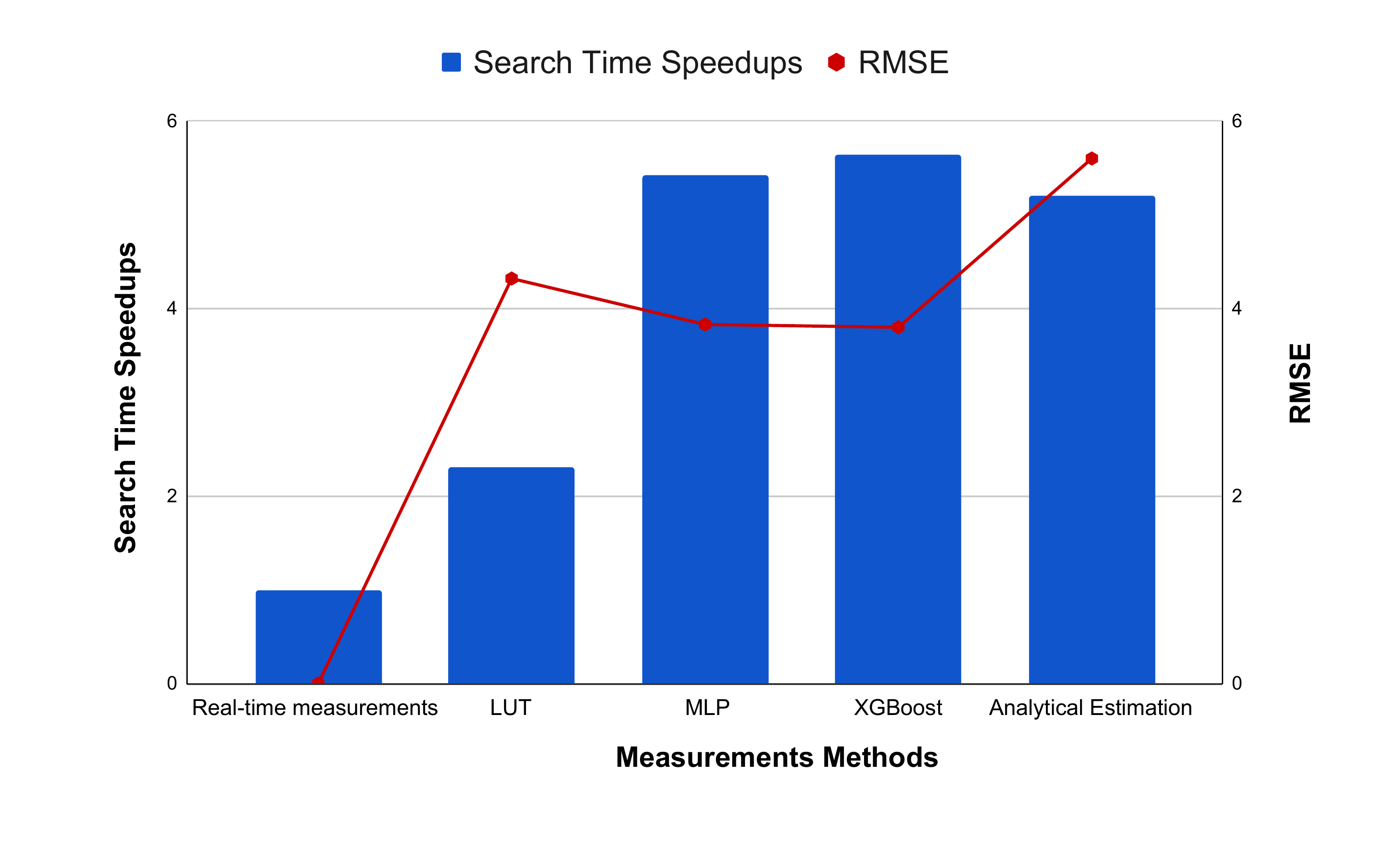}
    \caption{Comparison of hardware cost measurement methods. LUT stands for Look Up Table. The speedups are calculated w.r.t the real-time measurements. The exact statistics are displayed in table \ref{tab:hw_cost_comp}.}
    \label{fig:hw_cost_comp}
\end{figure}

Table \ref{tab:cost} summarizes all the methods used to estimate the latency, energy consumption in different NAS methods. 
 
\begin{table*}[!t]
    \centering
    \begin{tabular}{|p{3cm}|p{5cm}|p{3cm}|p{3cm}|}
    \hline
    Method & How the method is achieved ? & Hardware Cost Metric &  References \\ 
    \hline
    \multirow{2}{3cm}{Real-time measurements}  & \multirow{2}{=}{The sampled model is executed on the hardware target while searching.} & Latency & \parbox[t]{3cm}{MNASNet \cite{tan2019mnasnet} \\ NetAdapt \cite{netadapt} \\ Z.  Guo et al. \cite{guo2019single} \\ MCUNet \cite{lin2020mcunet}} \\ \cline{3-4} 
    && Energy  & \parbox[t]{3cm}{NetAdapt \cite{netadapt} \\MONAS \cite{hsu2018monas}\\ C. Gong et al. \cite{gong2019mixed}}\\ \hline
    Lookup Table Models & A lookup table is created beforehand and filled with each operator latency on the targeted hardware. Once the search starts, the system will calculate the overall cost from the lookup table. &Latency &  FBNet \cite{wu2019fbnet}
    HotNAS \cite{jiang2020standing}\\
    \hline 
    \multirow{4}{3cm}{Analytical Estimation}  & \multirow{4}{=}{Compute a rough estimate using the processing time, the stall time, and the starting time.} & Latency  & \parbox[t]{3cm}{FNAS \cite{FPGA} \\ NASCaps \cite{marchisio2020nascaps}\\ A. Anderson et al. \cite{performance-oriented} \\  Q. Lu et al. \cite{co-exploration20192}}\\ \cline{3-4} 
    && Energy & NASCaps \cite{marchisio2020nascaps}  \\ \cline{3-4}
    && Memory footprint & NASCaps \cite{marchisio2020nascaps} \\\cline{3-4}
    && Area & NASAIC \cite{NASICS}\\
    \hline 
    \multirow{3}{3cm}{Prediction Model} & \multirow{3}{=}{Build a ML model to predict the cost using architecture and dataset features.}   & Latency & proxylessNAS \cite{cai2019proxylessnas} NASAIC \cite{NASICS}
    NeuNets \cite{sood2019neunets} LEMONADE \cite{hsu2018monas}\\
    \hline
    \end{tabular}
    \caption{Summary of Hardware Cost Estimation Methods }
    \label{tab:cost}
\end{table*}

Real-world measurements provides a high accuracy in measuring the hardware efficiency of  an  architecture.  
MnasNet \cite{tan2019mnasnet} uses this method in the exploration. It achieves 75.2\% top-1 accuracy with 78ms latency on a Pixel phone platform, which is 1.8x faster than MobileNetV2 with 0.5\% higher accuracy. However, this method considerably slows down the search algorithm by averaging hundreds of runs to get precise measurements. 
Additionally, this strategy is not scalable and requires that all the hardware platforms are available. This solution could be costly and needs a lot of mobile devices and software engineering work. That's why many works tend to use a prediction model \cite{cai2019proxylessnas, zhang2020fast, co-exploration2019, co-exploration20192, FPGA, hao2019fpgadnn} or a pre-collected lookup table \cite{wu2019fbnet, abdelfattah2020best, jiang2020standing} or computing an analytical estimation \cite{FPGA, marchisio2020nascaps}. 

In ProxylessNAS \cite{cai2019proxylessnas}, the authors have developed three latency prediction models for three different platforms: a mobile phone (Google Pixel 1), a GPU (NVIDIA V100) and a CPU (Intel Xeon). To build their mobile latency predictors, they use the type of operators, input and output feature map sizes and other architectural hyperparameters as features. The real values for Pixel 1 phone have been measured with Tensorflow-Lite as software. On ImageNet, their model achieves 3.1\% better top-1 accuracy than MobileNetV2, while being 1.2x faster with measured GPU latency.

In NASCaps \cite{marchisio2020nascaps}, the functional behavior of a given specialized CNN and CapsNet hardware accelerator is modeled at a high level, to quickly estimate the memory usage, energy consumption, and latency.  The HW platform, the ASIC CapsAcc in the paper, is described at the RTL-level. Using a VLSI CAD tool and the RTL specifications, energy, memory, latency costs of elementary operations are measured. Elementary operations' cost correspond for example to the number of cycles and energy required to execute a layer. These values are then multiplied by the number of occurrences of each operation in the architecture and accumulated to obtain the total cost for the latency and energy. 

Although these techniques are efficient, they require hardware experts to build the models. For the lookup table method, for example, the researcher needs to dedicate a lot of time to optimize the code of each operator/architecture in the targeted hardware, which requires compilation knowledge. Similarly, to build the best model predictor, the researcher needs expert knowledge to select the best features and verify the results. Therefore, these methods impose a barrier to non-hardware experts.

In order to fairly compare the accuracy of each method, we computed the latency of each architecture in NAS-Bench-101 \cite{ying2019nasbench101} and compare it to the real-time measurements, according to three collecting methods: lookup table, prediction model and analytical estimation. For the lookup table, we calculated the latency of each operator used in the cell of the benchmark including Identity, Conv3x3BnRelu, Conv1x1BnRelu, MaxPool3x3, BottleneckConv3x3, BottleneckConv5x5, and MaxPool3x3Conv1x1. When a cell is generated, we sum the latency of the constructing operators and we get the latency of the whole cell. For the prediction model, we used two different models: a simple MLP and XGBoost, both trained on the real-time measurements of a portion of the benchmark (training set). We choose these two methods because they are both used by popular HW-NAS in \cite{cai2019proxylessnas} and \cite{wu2019fbnet} respectively. Lastly, for the analytical estimation we computed the number of MAC for the cell and multiply that by the latency of one multiply-add tensor instruction. 

This experiment was run on a Tesla K80 GPU, the prediction model MLP had to run for 50 epochs with early stopping. NAS-Bench-101 defines more than 400k cells, we have done our test on 165,580 cells. The search algorithm used to calculate the search time is an evolutionary algorithm based on the validation accuracy given by the benchmark and the latency measured by the different methods. Table \ref{tab:hw_cost_comp} presents the results of the accuracy values of different methods. As expected, the analytical estimation does not produce good results compared to the prediction models or the lookup table method. The prediction models even with a simple XGBoost give the best results and accelerate the search more than 5 times compared to the real-time measurements.

\section{Other Considerations for Hardware-aware NAS}
{\label{Section:Other_Considerations}}
Prior and Parallel to the hardware-aware NAS efforts, researchers have been working on reducing the memory footprint of deep learning models and execution time to facilitate the efficient deployment and design hardware-friendly models. Two main methods have been used, namely handcrafting new operators that are more efficient such as separable convolutions \cite{Chollet_2017_CVPR}, grouped convolutions \cite{krizhevsky2012imagenet}, and applying deep learning optimizations such as quantization \cite{zhang2019adaptive} and pruning \cite{asif2019ensemble}. This latter method is automated by several NAS works to compress the model and make it possible to execute on different hardware accelerators. Moreover, reducing the number of learnable parameters makes the training faster. For instance, SAL \cite{bouklihacene} reduces the number of parameters of ResNet-56 from 1.22M to 0.36M without a big degradation of the model's accuracy; 0.6\% decrease.

In this section, we are going to define the two most used deep learning optimizations and review the NAS works that focus on searching the right optimization parameters. 

\subsection{Automatic Mixed-Precision Quantization} \label{section:quantization} 
In the deep learning compression field, quantization is one of the most important methods. Starting with BinaryConnect  \cite{courbariaux2016binaryconnect} in 2015 to binarize CNNs weights, it is now implemented in many deep learning software such as PyTorch or Tensorflow. The idea is that the weights of the deep learning model do not have to be represented in full 32-bits precision and can be represented in 8-bits precision, or even binary precision in some cases, without significantly decreasing the model's accuracy. This idea was extended to the activations and weights \cite{courbariaux2016binarized}. Recently, mixed-precision quantization that applies different bitwidth values for different layers in the same network is more commonly used. 

\textbf{HAQ} \cite{han2019design} implemented a dedicated reinforcement learning agent that learns to assign the right bitwidth to each layer. Their goal is to specialize the architecture to a specific hardware platform by incorporating hardware constraints and accuracy into the reward function. For each layer, the agent takes two decisions: one for the weights and another one for the activation. 

\textbf{Single-Path} \cite{guo2019single} proposes an evolutionary algorithm that searches for the mixed-precision quantization policy. But the search costs a huge amount of time and processing as the space for mixed-precision is enormous. 

\textbf{BP-NAS} \cite{yu2020search} cast the mixed-precision quantization problem as a constrained optimization. In addition, it proposes a new constraint that encourages the model to focus on valid architectures while imposing a large punishment to the quantization outside the valid domain.

Although, these works present interesting solutions to the mixed-precision problem, the huge search space and computational cost of the learning process still represent a real challenge. Furthermore, mixed-precision representation requires specific MAC architectures with scalable-precision. This places a cap on the power efficiency, according to this study \cite{MAC_architectures}. With this in mind, \cite{anonymous2021uniformprecision} proposes a uniform quantization search algorithm called neural channel expansion (NCE). NCE expands the number of channel of a layer when it is more sensitive to the quantization error while maintaining the same precision level.

\subsection{Automatic Pruning}\label{section:pruning} 
The second prominent compression method is pruning. Pruning methods eliminate some neurons or connections according to a defined criterion to reduce the number of parameters in the architecture. Generally, we evaluate the importance of the neurons, we prune the least important ones, and finally, we fine-tune the network. 

\textbf{AMC} \cite{han2019design} proposes an automatic model compression that looks for the optimal sparsity for each layer during pruning. The authors trained a reinforcement learning agent to predict the best sparsity for given hardware. The reward function includes accuracy and FLOPs after pruning the architecture. 

X. Dong  and  Y.  Yang in \cite{dong2019network} propose to prune the over-parameterized network without performance damage. They directly search within their NAS process for a network with a flexible channel and layer sizes. \textbf{ABCPruner} \cite{lin2020channel} uses artificial bee colony algorithm to efficiently find the optimal pruned structure. Another worth mentioning paper is \textbf{Partial Order Pruning} \cite{Li_2019_CVPR} which proposes a hardware-aware NAS that prunes the search space with a partial order assumption to look for the best speed and accuracy trade-off.

\subsection{Security and Reliability Considerations} \label{section:security} 
Other NAS methods try to address safety-critical issues by discovering architectures that are robust against adversarial attacks \cite{robust, cubuk2017intriguing}. RAS \cite{robust} formulates the robustness as the sum of the accuracies on a bunch of adversarial samples. This robust evaluation makes it easier for the evolutionary algorithm to select better architecture in the population and apply different mutations (e.g., add a block, remove a block, add a connection...). Along with \cite{robust2}, they reveal these observations: first, the more dense the architecture is the more robust it is, second, under  computational  budget, adding  convolution  operations  to  direct  connection  edge is effective, and finally, flow of solution procedure (FSP) matrix is a good indicator of network robustness. Please note  here, that none of the NAS methods that consider security and reliability in the search place are hardware-aware.

\section{Industrial Adoption of NAS}\label{section:industry} 
One salient aspect that is not discussed in most AutoML surveys is its applications in the industrial domain. In this section, we compare different tools that apply NAS principles in order to let users build a specific model for their datasets.

\begin{enumerate}
    \item {\bf{Auto-Keras}} is a framework that is based on the deep learning framework \textit{keras}. By far, auto-keras is the most used autoML tool with features allowing to explore a wide range of neural network types (CNN, RNN and MLP). The search runs in parallel on CPUs and GPUs, with an adaptive search strategy for different GPU memory limits. Auto-keras builds its search space by applying a set of network morphism operations which keep the functionality of a neural network while changing its neural architecture. 
    
    \item \textbf{Microsoft NNI} (Neural Network Intelligence) is a general AutoML toolkit to help users tune their machine learning models (e.g., hyperparameters), design neural network architecture. One property it focuses on is the efficiency of their automation. For example, it leverages early feedback to speedup the tuning procedure. The proposed NAS framework helps the user define their own super network. For example, one can specify multiple operators for one single layer including depthwise convolution, dilated convolution, maxpooling, and the NNI will automatically find the best candidate based on a gradient-based optimization. On the other hand, the toolkit is also customizable for NAS researchers and enable them to modify the search algorithm and/or objective function. 
    
    \item {\bf{Google AutoML}} The idea with Google AutoML is to create a meta-model capable of learning a way to design, generate and propose architectures for a given dataset. The model will eventually achieve an accuracy on the data set that will be used as a reward to guide the reinforcement learning agent to explore and find the best architecture from the search space. Google AutoML generalize the use of machine learning to everyone by offering an easy-to-use toolkit to upload the data set and let the system find the appropriate model. The AutoML toolkit includes AutoML Vision that manipulates image data set and tasks and AutoML NLP that processes text datasets and focuses on the translation task. 
    
    \item {\bf{IBM NeuNets}} NeuNetS runs on the IBM cloud environment. The framework is designed to minimize human interaction and automate the whole machine learning pipeline, from the pre-processing methods to the algorithm search and model deployment. The process involves three components TAPAS, NCEvolve and HDMS. TAPAS synthesizes the neural network by offering a smart accuracy predictor based on the dataset characterizations and architecture features. NCEvolve searches for top-performing networks by minimizing the training time and resource needs. HDMS uses reinforcement learning to find and tune the architecture tailored for less common datasets. Finally, NeuNetS framework applies an evolutionary algorithm to tune some architectural hyperparameters such as the convolution filters. 
    
    \item {\bf{TPOT}} \cite{tpot} (Tree-based Pipeline Optimization Tool) is designed as a Python package that optimizes the machine learning pipeline using genetic programming. However, as it is based on scikit-learn framework, the package only allows classical machine learning algorithms such as random forest or decision trees and small MLPs and focuses only on tabular data sets. The package is still under development and may include more data types in the future.
    
    \item \textbf{Microsoft Archai} Archai is a platform designed exclusively for Neural Architecture Search. It allows NAS researchers to easily mix and match between different techniques while ensuring reproducibility and fair comparison between the search algorithms and speedup techniques. The framework also considers mutli-objective search optimizations by providing reports on model statistics such as number of parameters, flops, inference latency and model memory usage. 
    
    \item \textbf{deci.ai AutoNAC} deci.ai is a startup that aims to accelerate deep neural network inference on any hardware while preserving accuracy. AutoNAC is one of their commercial tools. As inputs, the framework needs a training and testing dataset, an already trained model and access to the hardware platform over which the model should be deployed. AutoNAC will then automatically compute a new low-latency model that preserves the accuracy. The idea behind AutoNAC is to construct small and fast models such that each model specializes in a small portion of the dataset, the basic strategy uses divide-and-conquer paradigm, where a complex problem is decomposed into a set of simpler sub-problems. For example, rather than building one complex model that classifies the data into the three classes defined by the data set, three simpler models that separate the classes two-by-two are built. 
    
    \item \textbf{Darwin} is a cloud-based service that utilizes genetic programming to iteratively select the best architectural hyperparameters and pass them to the next generation. Given a target dataset, Darwin tries to find a previously trained dataset that shares the same characteristics and initiate the search with the best model for that dataset. The framework also includes automation for other machine learning components such as the feature extraction by automatically detecting the data types and data pre-processing. 
\end{enumerate}

\begin{figure*}[!ht]
    \centering
    \includegraphics[trim={1cm 22.5cm 0 1.5cm},clip]{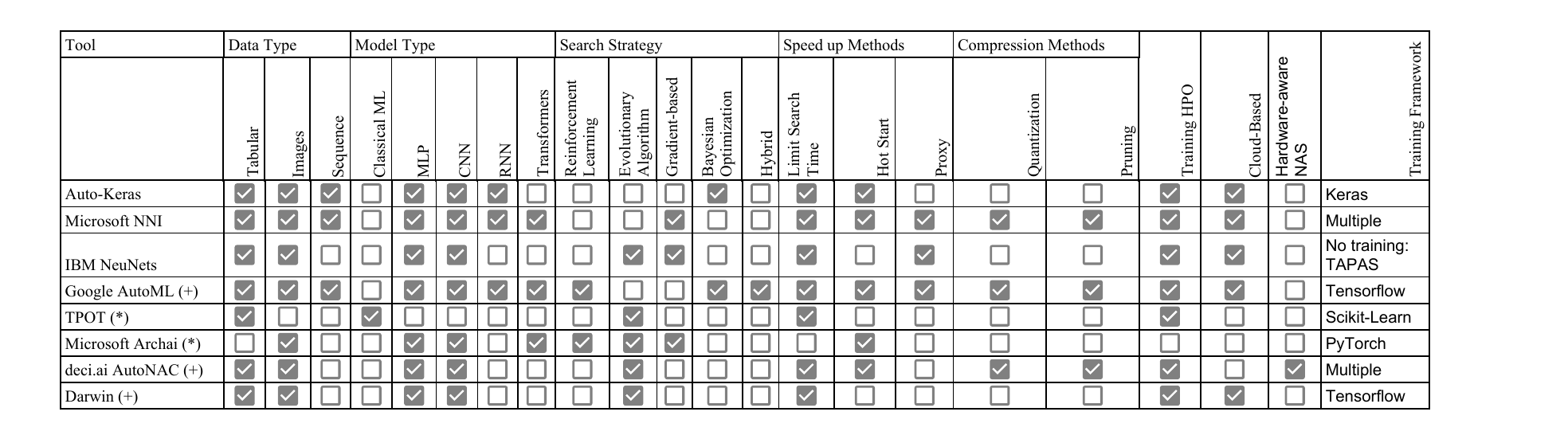}
    \caption{Comparison table of functionalities for NAS tools. (*) still under development, (+) commercialized tools. }
    \label{fig:tools}
\end{figure*}

Each of these tools have their own advantages and disadvantages as well as different spaces to compete with. Microsoft NNI, Archiai are designed to help NAS researchers compare, reproduce and experiment with multiple NAS search strategies and speedup methods. AutoKeras and TPOT are designed for the general machine learning researcher or engineer. Although they restrict their search space to a small set of architectural hyperparameters, the performance optimizations of their search are really attractive as an exchange for accuracy in this settings. In addition, both are open source which allow the user to run locally their search for privacy reasons.
On the other hand, IBM Neunets, Google AutoML and deci.ai AutoNAC try to demcratize make machine learning and make it available and accessible to everyone regardless of their skill level and without writing any piece of code. It is a more ambitious goal which is the reason why they are using more limited architecture search spaces. However, their search algorithms consume a lot of computing power and time that's why they are usually offered within a cloud service. The majority of the surveyed industrial NAS frameworks are not hardware-aware except for AutoNAC which given a defined model and its accuracy tries to find a faster architecture that doesn't reduce the specified precision. Nonetheless, Microsoft NNI and Google AutoML apply some compression techniques to the found architecture to specialize it in the target hardware.

\section{Challenges and Limitations}
{\label{Section:Challenges_Limitations}}
In this section, we lay down the main challenges and barriers that prevent us to unfold hardware-aware NAS's full potential. Most of the challenges are also applicable to general NAS methods. 

\subsection{Benchmarking and Reproducibility}\label{section:bench} 

An important challenge while working on NAS and HW-NAS is the reproducibility of the methods, which allows comparisons and concrete improvements. Unfortunately, due to the use of different search spaces, various training methods, and the required significant computational resources,  reproducibility is a difficult step. This difficulty is even higher when it comes to HW-NAS considering the numerous possibilities of targeted hardware devices. To address this issue, many works \cite{ying2019nasbench101,Dong2020NAS-Bench-201:,  Zela2020NAS-Bench-1Shot1:, klyuchnikov2020nasbenchnlp} have proposed different benchmarks and data sets that allow NAS researchers to: 
\begin{itemize}
    \item Eliminate the cost of generating a search space by querying directly a tabular data set. 
    \item Evaluate different search strategies on the same search space which allows a fair comparison between them. 
    \item Provide data sets to be used by accuracy predictor models and hardware cost models.
    \item Open HW-NAS research to non-hardware experts by proposing datasets that contain the hardware-related metrics. These metrics are usually obtained after optimization of the operators and software that require specific hardware knowledge. 
\end{itemize}

In this section, we will review each NAS benchmark and highlight their strengths and weaknesses.\\ 

\begin{table*}[!ht]
    \centering
    \begin{tabular}{|p{1cm}|p{2.5cm}|p{1cm}|p{1cm}|p{1cm}|p{1cm}|p{1cm}|p{1cm}|p{2cm}|}
         \hline 
         \multicolumn{2}{|c|}{}& NAS-Bench-101 \cite{ying2019nasbench101}& NAS-Bench-201 \cite{Dong2020NAS-Bench-201:}& NATS-Bench \cite{dong2020nats} & NAS-Bench-1shot1 \cite{Zela2020NAS-Bench-1Shot1:}& NAS-Bench-NLP \cite{klyuchnikov2020nasbenchnlp}& NAS-Bench-301\cite{anonymous2021nasbench} (DARTS) & HW-NAS-Bench \cite{li2021hwnasbench}\\
         \hline \hline
         \multicolumn{2}{|l|}{Arch Search Space}  & Cell-based CNN & Cell-based CNN & Cell-based CNN& Super Network CNN & Cell-based LSTM  & Super Network CNN  & Cell-based CNN (NAS-Bench-201 + FBNet) \\
         \hline 
         \multicolumn{2}{|l|}{Size (number of architectures) } & 423k & 15,625 & 15,625   & 363,648 & 14k & $10^{18}$ & $46875 + 10^{21}$ \\
         \hline
         \multirow{2}{=}{Datasets} & CIFAR-10   & \checkmark & \checkmark &\checkmark  & \checkmark & & \checkmark& \checkmark \\ \cline{2-9}
         & CIFAR-100 & \checkmark & \checkmark &\checkmark & \checkmark &  & &  \checkmark \\ \cline{2-9}
         & ImageNet &  &  &\checkmark   & \checkmark & & & \checkmark \\ \cline{2-9}
         & PTB &  &  &   &  & \checkmark&  &  \\ \cline{2-9}
         & WikiText2 &  &  &   &  & \checkmark &&  \\ \hline
         \multirow{6}{=}{Metrics} & Validation Accuracy   & \checkmark & \checkmark &\checkmark   &\checkmark &\checkmark&\checkmark&  \checkmark \\
         \cline{2-9} 
         & Training Time & \checkmark&  \checkmark & \checkmark &\checkmark & \checkmark&\checkmark & \checkmark \\
         \cline{2-9}
         & Trained Parameters  & \checkmark&  \checkmark&\checkmark  & &\checkmark & &\checkmark  \\
         \cline{2-9}
         & FLOPs  &  & \checkmark &\checkmark & &&  &\checkmark  \\
         \cline{2-9} 
         & Test Accuracy  & \checkmark&  \checkmark& \checkmark &\checkmark &\checkmark &\checkmark &  \checkmark \\
         \cline{2-9} 
         & Latency  & & \checkmark &\checkmark  &\checkmark & & \checkmark &\checkmark \\
         \cline{2-9} 
         & Energy & &  &  & & && \checkmark  \\
         \hline 
         \multicolumn{2}{|l|}{Predictor Models} &  &  &  & & & \checkmark &\checkmark \\ \hline
         \multirow{4}{=}{Hw Platforms} & GPU  & GTX 1080Ti &GTX 1080Ti & Not Mentioned& Not mentioned & Tesla V100-SXM2 & Not Mentioned  & Edge GPU Jetson TX2 \\
         \cline{2-9} 
         & Edge TPU  & & &  &  &  &  & Edge TPU Dev Board  \\
         \cline{2-9} 
         & Smartphone  & & & &  &  &  & Pixel 3 \\
         \cline{2-9} 
         & Raspberry Pi  & & & &  &  &  & Raspi 4 \\
         \cline{2-9} 
         & FPGA  & & & &  &  &  & Xilinx ZC706 \\
         \cline{2-9} 
         & ASICs  && & &  &  &  & ASIC-Eyeriss \\
         \hline 
    \end{tabular}
    \caption{Comparison of NAS Benchmarks.}
    \label{tab:my_label}
\end{table*}

\textbf{NAS-Bench-101} \cite{ying2019nasbench101}\footnote{Open sourced at https://github.com/google-research/nasbench} is a tabular dataset that maps 432k unique architectures to their relative training accuracy, validation accuracy, testing accuracy as well as training time and the number of trained parameters. Each architecture is trained for various numbers of epochs {4, 12, 36, 108}. The architectures follow a fixed macro architecture, with searchable cells stacked (See section \ref{section:arch_ss}). Each cell is constructed with up to 7 layers that can include 3 types of operations: 3x3 conv, 1x1 conv, and 3x3 max-pooling. NAS-Bench-101 allows flexibility by allowing different layers to be used in the stacked cells. The search space is constrained by limiting the number of edges to 9 and the number of nodes to 7 including the input and output nodes for each cell.  However, not including operations such as separable convolution or dilated convolutions makes the resulting models parameter-heavy, which is not hardware-friendly. Another downside of this benchmark is the inability to use one-shot optimizer NAS, this is due to the tabular format.
To enable the evaluation of weight-sharing NAS methods, two benchmarks have been released NAS-Bench-201 \cite{Dong2020NAS-Bench-201:}\footnote{Open sourced at https://github.com/D-X-Y/AutoDL-Projects} and NAS-Bench-1Shot1 \cite{Zela2020NAS-Bench-1Shot1:}\footnote{Open sourced at https://github.com/automl/nasbench-1shot1}.

\textbf{NAS-Bench-201} represents 15,625 architectures using a fixed cell-based macro architecture. Similar to NAS-Bench-101, it uses a predefined set of operations including conv 3x3, conv 1x1, Avg pooling, skip connection and no operation label. Each architecture is trained on three different datasets with different complexities namely, CIFAR-10, CIFAR-100 and imagenet-16. The authors extended this benchmark and presented a year later NATS-Bench. \textbf{NATS-Bench} \cite{dong2020nats} increased the search space size by varying the sizes of their architectures. 

Similarly, \textbf{NAS-Bench-1Shot1} presented a new reformulation to reuse the even much more extensive computation of the NAS-Bench-101 dataset (120 TPU years) to create three new one-shot search spaces with growing complexity containing 6240, 29160, and 363648 architectures. In order to use the expensive experimentation done by NAS-Bench-101, the authors created a one-shot architecture that contains all the discrete cell architectures defined by NAS-Bench-101. This allows the search algorithm to only train the supernetwork and get the performance of each path from the NAS-Bench-101. Therefore, their benchmark construction does not need any additional cost. 

Alternatively, \textbf{NAS-Bench-NLP} defines a tabular benchmark for NLP tasks. Their cell-based search space is constructed based on an LSTM Macro-architecture borrowed from AWD-LSTM \cite{awd-lstm}. Each cell can contain up to 24 nodes with 3 hidden states and 3 linear input vectors. With these constraints and a set of operators composed by the most used activations in recurrent cells, they are able to build LSTM, GRU cells and many more. The architectures are trained on Penn Tree Bank (PTB) dataset \cite{ptb} and a sub-sample of the best performing ones are also trained on WikiText-2 \cite{wikitext}, which is a more realistic dataset for real-world NLP problems. 

\begin{figure}[!ht]
    \centering
    \includegraphics[width=9cm]{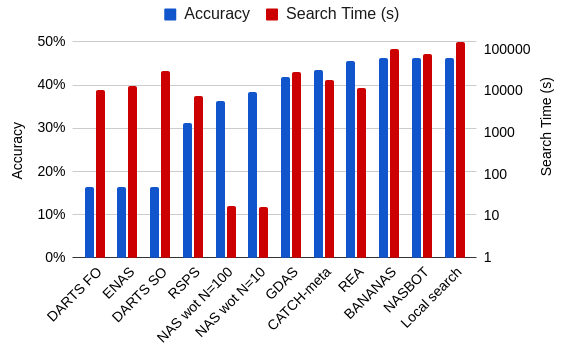}
    \caption{Results of different search algorithms on NAS-Bench-201.}
    \label{fig:bench}
\end{figure}

A major disadvantage of these benchmarks is the size and diversity of their search spaces. Indeed, as presented by the experimentation results we obtained in figure \ref{fig:bench}, on the small dataset NAS-Bench-201, a local search, which is the simplest optimization strategy, achieves state of the art results without a significant search time compared to other search strategies, except NAS without training \cite{naswot}. In this experiment, we compare the different results obtained by different NAS approaches on the NAS-Bench-201. Most of them use the metrics provided by the benchmark along with fine-tuning the architecture to obtain a more accurate validation metric, except NAS without training \cite{naswot}.  The NAS searche without any further training achieves decent results within 17sec of search. By dividing the benchmarks into $N$ mini-batches, they increase their training efficacy. The higher this number is ($N$), the higher the over-fitting probability on the benchmark. Therefore, using small datasets with complex search algorithms does not yield any good results in terms of accuracy of the final architecture or efficiency of the search.

\textbf{NAS-Bench-301} is a much bigger benchmark that was designed to overcome the over-fitting problem on the architectures. It is based on DARTS \cite{liu2019darts} search space. Since DARTS search space is far too large to be exhaustively evaluated by real-world measurements, the authors built a surrogate model capable of predicting the various performance metrics. This model is trained on a subsample of the benchmark, 60k architectures whose performance have been measured on CPUs.

A more recent paper introduced the first hardware-aware NAS benchmark, \textbf{HW-NAS-Bench} \cite{li2021hwnasbench}. This work extends the number of hardware metrics and records the latency, the energy consumed on 6 hardware devices including commercial edge devices, FPGA, and ASIC . Its search space is a combination of FBNet \cite{wu2019fbnet} search space and NAS-Bench-201. 

Although these datasets provide a good start to test different search strategy, they lack a lot of important operators that can significantly change the resulting architecture for hardware-aware NAS. In addition, we note a growing number of different applications of NAS in various tasks such as image restoration \cite{EAIR, zhang2019ir, gong2019autogan, ho2020neural}, semantic segmentation \cite{liu2019auto, shaw2019squeezenas}, and medical segmentation \cite{weng2019unet, zhu2019v}. Therefore, there is a growing need for proper benchmarks for each of the diverse tasks.

\subsection{Transferability of the AI Models}\label{section:transfer} 

A concept that is quite popular in deep learning is transferability. We train the model on a proxy dataset; that is usually smaller and simpler, then we fine-tune the trained weights on the targeted dataset to get a more efficient model. For example, we can train our image classifier on CIFAR-10 \cite{zoph2018learning}, which contains 10 classes of 32x32 pixels images, and then fine-tune our model on ImageNet \cite{russakovsky2015imagenet}, which is made of 14 million images and 20,000 classes. Another benefit of transferability is when the target dataset is not large enough to train a model on. 

To enhance transferability, previous NAS works used cell-based search spaces. To transfer a model obtained from a cell-based search spaces, we just need to adjust the input sizes of the cells and deepen the network by adding more cells. However, stacking the same blocks seems to be not efficient when incorporating hardware constraints. As MNASNet \cite{tan2019mnasnet} argued, restricting the cell diversity is critical for achieving high accuracy and low latency on mobile settings. Therefore, MNASNet uses a hierarchical search space that diversify the cells in the architecture as well as the operators within that cell. 

Many NAS works \cite{liu2018progressive, nayman2019xnas} have included dedicated evaluations of the transferability of their final model. PNAS \cite{liu2018progressive} proved that CIFAR-10 classification is highly correlated to ImageNet classification, hence transferring a cell or an architecture from CIFAR-10 to ImageNet would produce good results. XNAS \cite{nayman2019xnas} transferred their final cell structure on 6 popular classification benchmarks surpassing other conventional NAS methods while taking account of the hardware constraints.

Another interesting concept is presented by NAT \cite{lu2020neural}. They leverage NAS process to directly find transferable weights (i.e., get rid of the fine-tuning stage). The key idea behind NAT is that they start from a supernetwork and adaptively modify it to obtain a task-specific super network. This latter approach can then be used directly to search for architectures within one task without the additional training cost. Under mobile scenarios, they demonstrated the efficacy of NAT on 11 benchmarks including ImageNet. 

Overall, taking the transferability of the model into account remains a difficult challenge. Most solutions modify the search space to fit a certain task by leveraging a task-specific macro-architecture. Hardware-awareness adds another level to this challenge as the architecture needs to be flexible to adapt to multiple platforms. Note that, transfer learning generalizes the ability to use a model from one task to another. In this section, we only talked about NAS that transfers their models from one dataset to another within the same task, which is a sub definition of transfer learning. Unfortunately, as NAS tasks evolve around image classification mainly, the whole search needs to be executed all over again if we want an architecture for another task.  

\subsection{Transferability of the Hardware-aware NAS across multiple Platforms}\label{section:hw_transfer} 
HW-NAS suffers from the conditional optimality due to the variety of existing devices. Ideally, we should design different architectures for different platforms. However in real situations, given the prohibitive cost of the search and the cost of training on multiple architectures, we often resort to designing one architecture and deploying it anywhere. Transferring a model from one platform to another or being able to produce hardware transferable models via the NAS process is an interesting challenge for HW-NAS. The main difficulty lies in the variety and complexity of the existing platforms. For example, different platforms might perform well on different types of convolutions. In the following section, we discuss two approaches. The first transfers a single-target HW-NAS to another target by modifying the measurement values. The second takes the final architecture of the NAS process and transforms it to fit another platform. Each approach has its pros and cons as discussed below.

\paragraph{Transfer the entire NAS process} this means that the whole NAS process needs to be re-executed to suit the new targeted hardware. In a single target HW-NAS, it is quite a challenge to transfer the entire process to another platform. Without mentioning the huge computational cost of retraining the entire process, the collection of the hardware constraints can be costly as well. When using real-world measurements, \cite{han2019design} ran the NAS search for three hardware platforms: CPU (Xeon E5-2640 v4), GPU (Tesla V100) and mobile phone (Google Pixel-1). However, using real-world measurements considerably slows down the search algorithm and requires the availability of the targeted hardware during the search time. On the other hand, using an analytical estimation requires expert knowledge for the different targeted platforms. When using other collection methods such as the lookup table or the prediction model, we'll need to collect data from the new platforms by running again the entire set of operators. To this end, HW-NAS tries to create a general measurement method. For example, Once-for-all \cite{cai2020onceforall} created a lookup table with the reported inference latency on each tested hardware platform (i.e., Samsung S7 Edge, Note8, Note10, Google Pixel1, Pixel2,LG G8, NVIDIA 1080Ti, V100 GPUs, Jetson TX2, Intel Xeon CPU, Xilinx ZU9EG, and ZU3EGFPGAs). According to the used measurement method, transferring the NAS process to target another platform is increasingly difficult and not scalable. 

\paragraph{Transfer the final model} An alternative approach is to find the best model for one hardware platform and then try to specialize it for another one. This was done by \cite{cai2020onceforall, cai2019proxylessnas, nayman2019xnas}. Usually, the specialization is done by compressing the model using quantization which enables the model to fit in tiny devices. However, the specialization is challenging because of the following reasons:
\begin{itemize}
\item \textbf{An operator may be efficient for one platform and less efficient in another:} In \cite{chu2020discovering}, the authors argued that separable convolutions give great results when ran on GPUs but perform badly on CPUs. Additionally, it is common that deeper networks perform well on CPUs while wider ones perform well on GPUs because of the possible parallelization. Table \ref{tab:operators} demonstrates the comparison between the execution times of different operators on an Intel i7 CPU and NVIDIA TX2 GPU . The results reinforce our assumption that different operators's efficiency vary from one platform to another. Therefore, the best model is highly correlated to the platform we choose which makes design a HW-NAS that targets multiple platforms very challenging. 

\item \textbf{Limits of the compression methods:} We consider here the quantization and pruning. For these two methods, we know that theoretically the compression ratio have a threshold that cannot be surpassed. For example, quantizing a model implies encoding its activations and weights into the minimum possible bit length. Theoretically this length is 2 (one bit). Even without considering the trade-off between accuracy and model size, these methods have limits. This is why we need to start from a model that is already optimized to be able to deploy the model on tiny devices. For this end, \cite{han2019design} is composed of three components. The first components is a multi-objective NAS that searches for the best model in terms of accuracy and latency. Given the resultant model of the first component, the second component searches for pruning possibilities that would preserve the accuracy and decrease the model size. Finally, the last component takes care of quantizing the model with a mixed precision. \cite{uniform} starts from a standard architectures such as VGG, ResNet and GoogleNet and casts the quantization as a neural architecture search problem. This work achieves a minimal loss of accuracy with appreciable memory savings. In addition to the limit for the model size, we also have a limit for the accuracy. The compressed model typically does not have a better accuracy than the pre-trained model we started from. 
    
\end{itemize}

\begin{table}[!ht]
    \centering
    \begin{tabular}{|p{2cm}||p{2cm}|p{2cm}|c|}
        \hline
         Operator & Avg Latency on CPU (ms) & Avg Latency on GPU (ms) & Accuracy\\
         \hline
         \hline
         Conv2d & 1.33 & 0.85 & 0.67  \\ 
         \hline
         Separable Depthwise Convolution & 2.05 &0.54& 0.64 \\ 
         \hline
         Dilated convolution & 1.36 & 0.835 & 0.56 \\ 
         \hline 
         Grouped Convolution &2.27& 1.94 & 0.62 \\ 
         \hline
         LSTM cell &14.93& 2.45& 0.57 \\
         \hline
         GRU cell & 9.32& 2.53 & 0.65 \\ 
         \hline
    \end{tabular}
    \caption{Comparison between different operators on Intel i7 CPU and  NVIDIA TX2 GPU. The convolution operators were used to create a CNN model that was trained for Image Classification on ImageNet. The RNN cells were trained on Text Classification on IMDB dataset \cite{imdb}. \textit{Results were obtained using PyTorch with number of samples 1000.}}
    \label{tab:operators}
\end{table}

\subsection{Outlook and Future Directions}\label{section:outlook}
In addition to the benchmarking, reproducibility and transferability challenges presented in the previous three sub-sections, HW-NAS suffers from the same limitations as NAS. The main one is the cost of the search algorithm. The current popular way to speed up the process is to use differentiable neural architecture search and create a supernetwork representing the whole search space. This supernetwork is trained once to get the weights of all the sub-networks, thus avoiding training each sampled architecture. This technique reduces the search time from several days to hours. However, a major disadvantage of this network is its restriction on the targeted task and domain. More research is needed to explore efficient ways and tricks to make HW-NAS more practical and more amenable to a diverse set of tasks and domains. This would especially be useful in commercial settings.

Given the increasingly fragmented hardware landscape, HW-NAS should investigate the possibilities to reduce the conditional optimality problem. Most existing NAS approaches design once and deploy to all. We have seen this clearly through the industrial adoption of NAS techniques which lacks hardware-awareness. Ideally, we should be able to design different architectures for different devices as no-one-model fits all. This problem is significantly exacerbated by the increasing hardware heterogeneity in AI-powered devices. 

There is no doubt that the future of mobile and handheld devices is AI. AI-focused mobile chips from top manufacturing companies like Apple, Samsung, Huwaei and others make their ways into the mainstream. These devices use SoCs that take advantage of multiple platforms (e.g., GPUs, CPUs and NPU in the same chip). However, this heterogeneity of platforms needs to be well understood to speed up the inference time of deep neural networks \cite{8963950}. This motivates further the research community to simultaneously explore both the architecture search space and the hardware design space to identify the best neural architecture and hardware pairs that maximize both test accuracy and hardware efficiency.  Such co-exploration will be critical to allow designing architectures that can be deployed efficiently on a variety of platforms: data center, edge, mobile, and embedded.

HW-NAS also suffers from a lack of benchmarks and public datasets that can help accelerate HW-NAS research and make it reproducible and accessible. HW-NAS-Bench \cite{li2021hwnasbench} is a recent first attempt towards this direction. It tries to offer computed metrics on multiple commercial platforms including edge GPUs, FPGA and ASICs and make these measurements available to the research community. One of their main goals is to democratize HW-NAS research to non-hardware experts.

More work is also needed to further co-explore neural architectures with quantization, pruning, and hardware design. Most of the existing efforts, such as mixed-precisions have been studied in the context of fixed architectures.

HW-NAS should also look at exploring neural architecture search with emerging computing paradigms such as in-memory-computing~\cite{9082822}. These new non-von-Neumann paradigms present novel solutions to AI computing based on emerging nano-devices called Phase-Change Memory (PCM)~\cite{8776518}. Existing HW-NAS solutions are dedicated to the conventional von-Neumann computing architecture, where the memory wall heavily limits performance. The optimization space spans multiple design points that range from device types, to circuit topologies, to device non-idealities and variations, to neural architectures.

While deep neural networks have clear commercial use cases, the next AI breakthrough may require an entirely different combination of algorithm, hardware and software. HW-NAS offers a paradigm that opens up the design space and pushes forward the Pareto frontier between hardware efficiency and model accuracy for efficient and improved hardware/software co-design, hence pushing AI to its next frontiers.


\section{Conclusion}
{\label{Section:Conclusion}}
We have provided a comprehensive survey and systematic analysis of state-of-the-art hardware-aware neural architectural search. We reviewed several multi-objective strategies that aim to find the optimal architecture with the highest accuracy while reducing memory and computations costs. We provide a HW-NAS taxonomy and categorize existing approaches along four key dimensions: the search space, the search strategy, the acceleration technique, and the hardware cost estimation strategy. We also discussed other considerations in Hardware-aware NAS that include optimizations such as quantization and pruning. Finally, we presented a discussion on future directions that would benefit existing and future HW-NAS researchers. We believe that HW-NAS is an important direction that would enable hardware-software co-design and make deep learning solutions not only accessible to everyone but also sustainable. 

\begin{landscape}
\begin{table}[!t]
    \centering
    \small
    \begin{tabular}{|c|p{2cm}|p{2cm}|c|p{2cm}|p{2cm}|p{2cm}|c|p{2cm}|c|c|c|}
        \hline 
         Reference & Target Model & Arch Space & HSS & Strategy & Hw Cost Estimation Method & Target Hw & HWT & Speedup Technique & P & THPO & OS   \\
         \hline
         \hline
         \cite{tan2019mnasnet} & Standard CNN & Factorized Hierarchical & &Reinforcement learning & Real-world measurements & Pixel 1 &  & early stopping &  \checkmark &  & \checkmark\\
         \hline
         \cite{cai2019proxylessnas} & Extended CNN & Layer-wise  & & Gradient-based 
        Optimization & ML Prediction Model & Pixel 1 &  & Super Network  & & \checkmark & \checkmark\\
        \hline 
         \cite{wu2019fbnet} & Extended CNN & Layer-wise &  & Gradient-based 
        Optimization & Lookup Table Model & Samsung S8 IPhone X & \checkmark & Super Network  & \checkmark &  & \checkmark\\
        \hline 
        \cite{lin2020mcunet} & Standard CNN & Optimized hyperparameters Search space &  & Evolutionary Algorithm & Try-to-deploy & MCU & & Super Network  & \checkmark & & \\
        \hline 
        \cite{FPGA} & Standard CNN & Hyper-parameters Search space & \checkmark & Reinforcement learning & Analytical Estimation & FPGA & \checkmark & Ignore not deployable models & \checkmark & \checkmark & \\
        \hline 
        \cite{NASICS}& Extended CNN & Hyper-parameters  & \checkmark & Reinforcement learning & MLPrediction Model & ASIC & \checkmark & Ignore not deployable models &  &  & \\
        \hline 
        \cite{marchisio2020nascaps} & Capsule + Standard CNN& Layer-wise  & \checkmark & Evolutionary algorithm & Analytical Estimation & ASIC & \checkmark & Early stopping & \checkmark &  & \checkmark\\
        \hline 
        \cite{chu2020discovering} & Standard CNN & Multi Hardware &  & Reinforcement Learning & ML Prediction Model & Multiple & \checkmark &  Super Network &  &  & \\
        \hline 
        \cite{jiang2020standing} & Extended CNN & Layer-wise  & \checkmark & reinforcement learning & Lookup Table Model & FPGA & \checkmark & Hot Start &  &  &  \\
        \hline 
        \cite{zhang2020fast} & Extended CNN & Optimized layer-wise  &  & Two stage One-Shot Search & ML Prediction Model & DSP /VPU/CPU & \checkmark & Super Network &  &  &\\
        \hline 
        \cite{FPNet} & Extended CNN & Hierarchical Search Space & \checkmark & Reinforcement Learning & Real-time measurements & FPGA &  & NOT Mentioned &  &  &  \\
        \hline 
        \cite{abdelfattah2020best} & Standard CNN & NASBench\cite{ying2019nasbench101} & \checkmark & Reinforcement Learning & Lookup Table Model & FPGA &  & NOT Mentioned &\checkmark&&\\
        \hline 
        \cite{hao2019fpgadnn} & Extended CNN &  Optimized layer-wise & \checkmark & Stochastic Coordinate Descent & ML Prediction Model & FPGA & \checkmark & Accuracy prediction & \checkmark &  & \\
        \hline 
        \cite{wang2020hat}  & Transformers & Encoder wise Search Space &  & Evolutionary Algorithm & MLPrediction Model & Raspberry-Pi/CPU/GPU & \checkmark & Super Network &  &  & \checkmark \\
        \hline
        
    \end{tabular}
    \caption{Detailed overview of 15 most popular HW-NAS. \hspace{\textwidth}
    HWT: Hardware transferability, P: proxy datasets, \hspace{\textwidth}
    THPO: training hyperparameter optimization, OS: Open source.\hspace{\textwidth}
    For a complete list of all HW-NAS works, please visit the following link \href{https://tinyurl.com/y6458skt}{https://tinyurl.com/y6458skt}
    }
    \label{tab:hw_nas}
\end{table}
\end{landscape}



%


\ifCLASSOPTIONcaptionsoff
  \newpage
\fi



%

%

\begin{IEEEbiography}[{\includegraphics[width=1in,height=1.25in,clip,keepaspectratio]{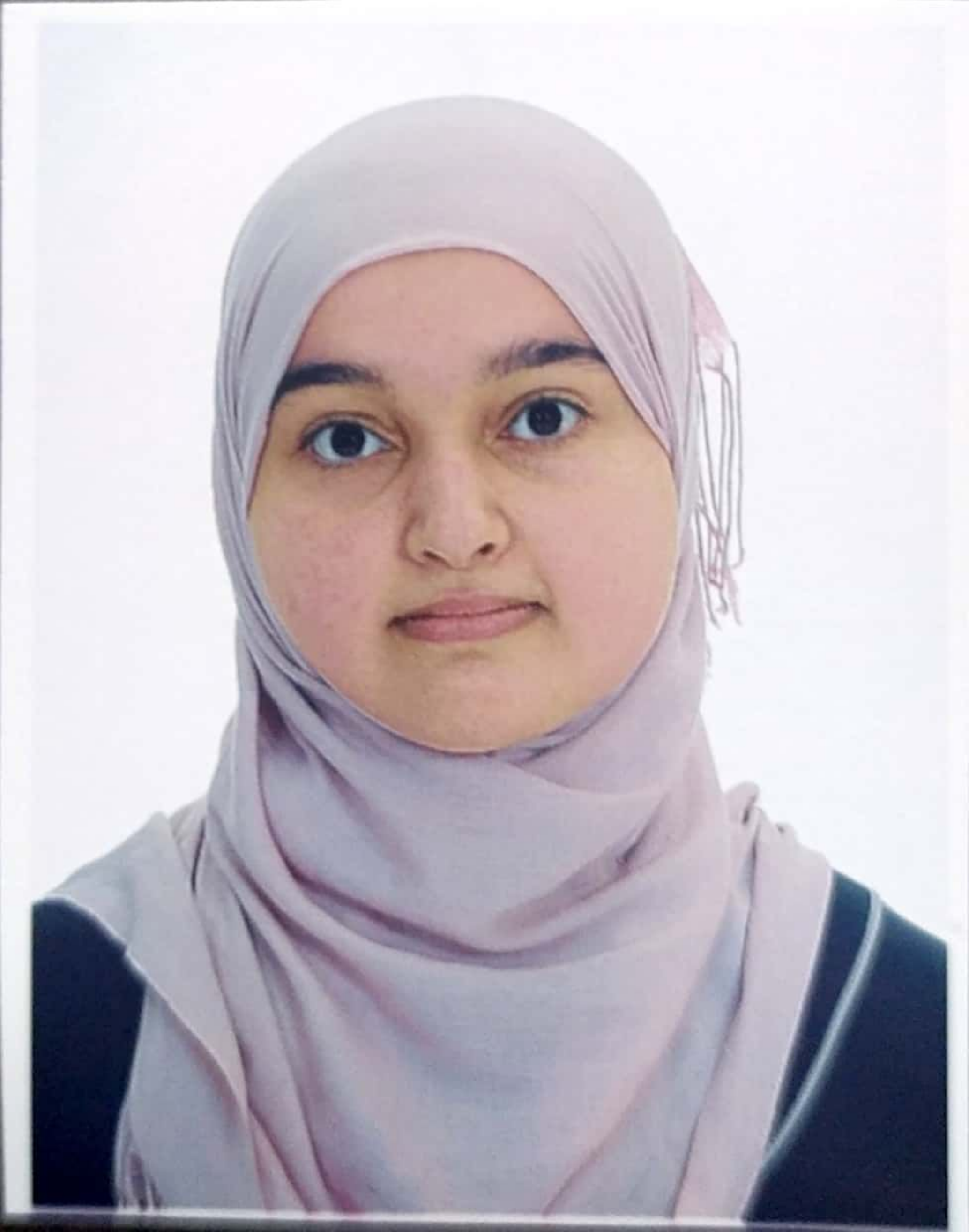}}]{Hadjer Benmeziane}
received her Master and Engineering degrees in Computer Science at the Higher National School of Computer Science, Algiers, Algeria in 2020. She is currently pursuing her PhD degree in Computer Science at LAMIH/CNRS, Universit\'{e} Polytechnique Hauts-de-France, Valenciennes, France. 
\end{IEEEbiography}


\begin{IEEEbiography}[{\includegraphics[width=1in,height=1.25in,clip,keepaspectratio]{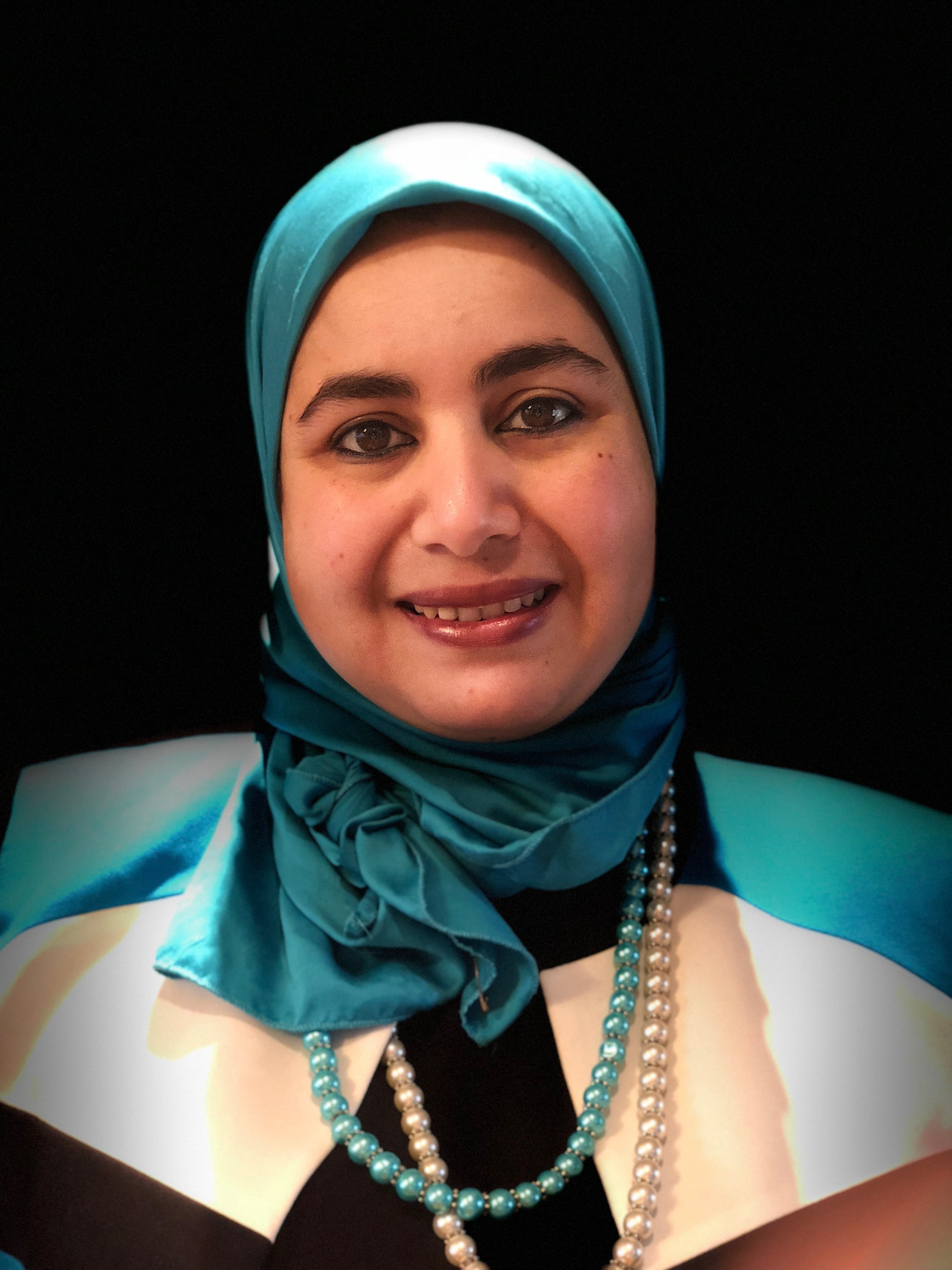}}]{Dr. Kaoutar El Maghraoui}
 received her PhD in computer science from Rensselaer Polytechnic Institute in 2007. Since then, she has been affiliated with the IBM T.J. Watson Research Center. Kaoutar currently holds the position of a principal research scientist at the IBM Research AI organization where she is focusing on innovations at the intersection of systems and artificial intelligence. She leads the End-Use experimental AI testbed of the IBM Research AI Hardware Center, a global research hub focusing on enabling next-generation chips and systems for AI workloads. Her research interests include AI platforms, distributed deep learning, neural architecture search, distributed systems, and AI hardware acceleration.
 
\end{IEEEbiography}

\begin{IEEEbiography}[{\includegraphics[width=1in,height=1.25in,clip,keepaspectratio]{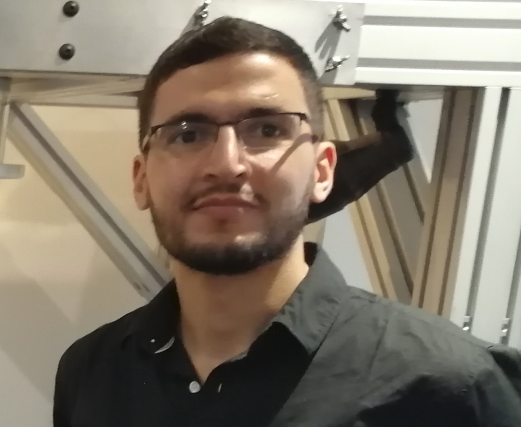}}]{Dr. Hamza Ouarnoughi}
received his PhD in computer science from the University of Western Brittany in 2017. 
 Since then, he has been associate professor at LAMIH/CNRS, Universit\'{e} Polytechnique Hauts-de-France and INSA Hauts-de-France, Valenciennes, France. His research interests include computer architecture and systems, cloud computing, storage systems, and artificial intelligence.
\end{IEEEbiography}

\begin{IEEEbiography}[{\includegraphics[width=1in,height=1.25in,clip,keepaspectratio]{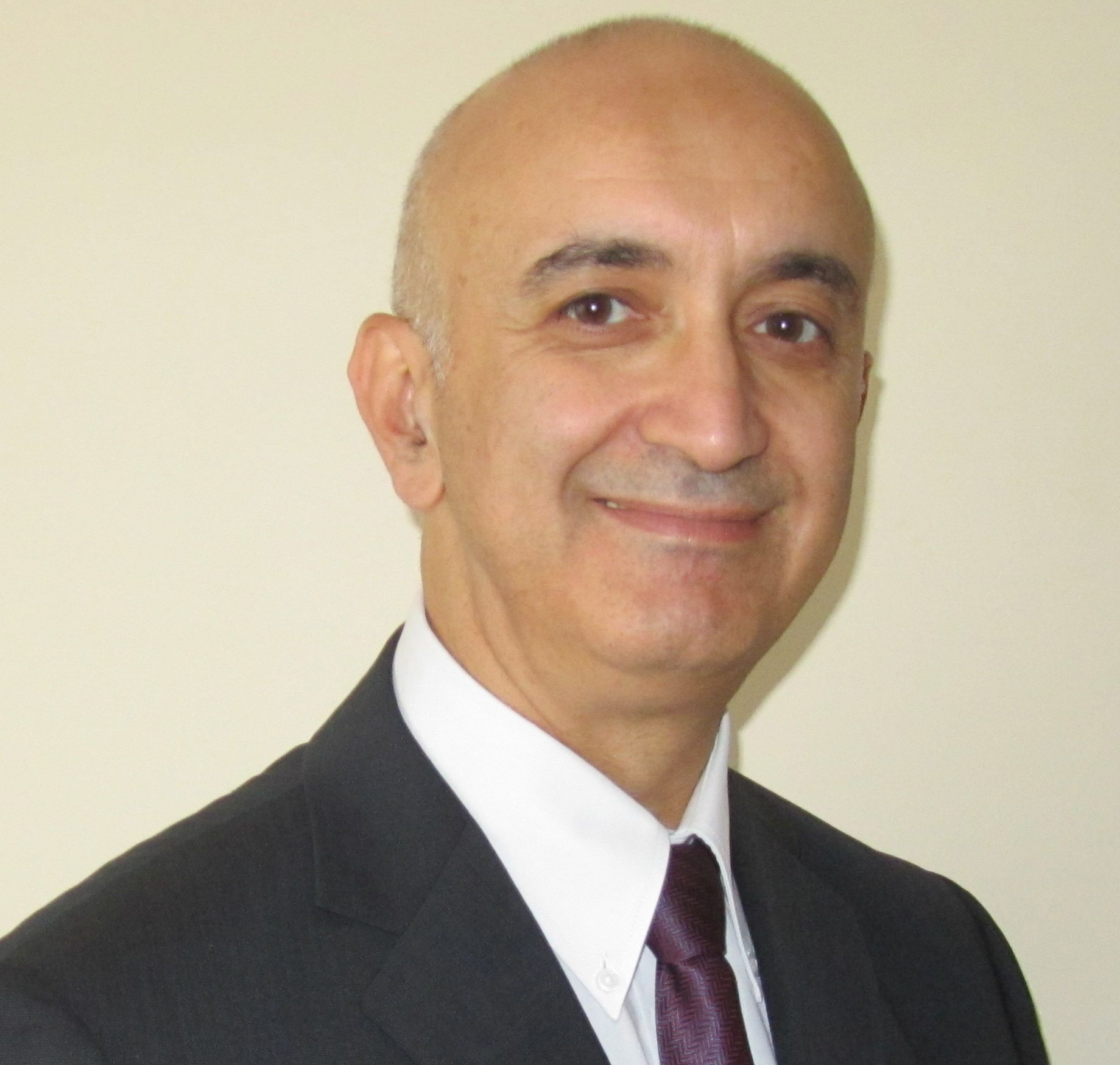}}]{Pr. Smail Niar} received his PhD in computer science from the University of Lille in 1990. Since then, he has been professor at LAMIH/CNRS, Universite Polytechnique Hauts-de-France and INSA Hauts-de-France, Valenciennes, France. He is member of the European Network of Excellence on “HIgh Performance and Embedded Architectures and Compilation” (HIPEAC), EuroMicro society and IEEE senior member. His research interests include heterogeneous multi-core architectures, autonomous driving and reliability for intelligent transportation systems. 
\end{IEEEbiography}

\begin{IEEEbiography}[{\includegraphics[width=1in,height=1.25in,clip,keepaspectratio]{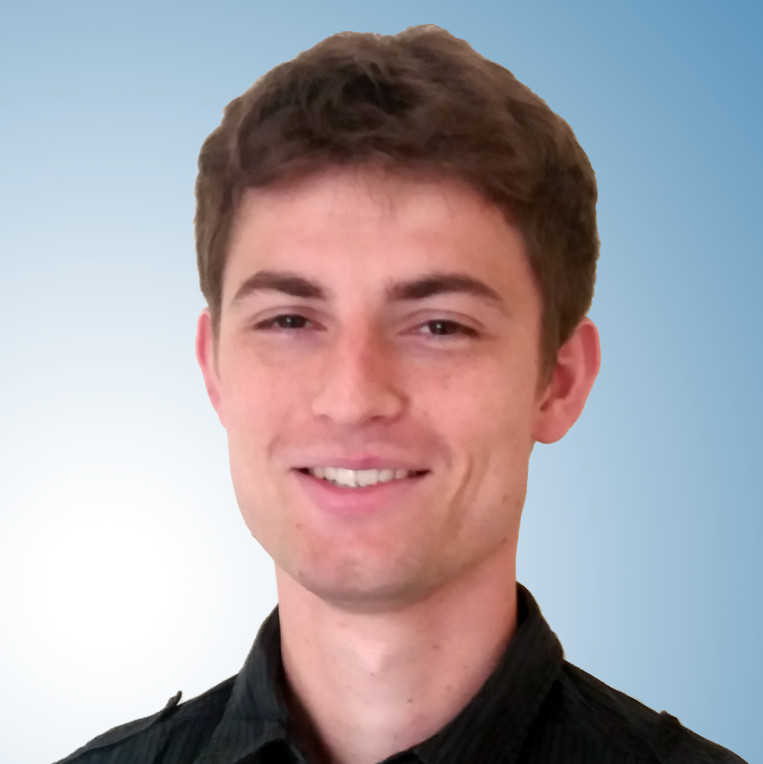}}]{Dr. Martin Wistuba}
is a researcher at IBM Research, where he develops tools to automate deep learning. Previously, he received his Ph.D. in Machine Learning from the University of Hildesheim. His research interest includes AutoML, in particular the idea of meta-knowledge transfer to speed up Bayesian optimization and Neural Architecture Search.
\end{IEEEbiography}

\begin{IEEEbiography}[{\includegraphics[width=1in,height=1.25in,clip,keepaspectratio]{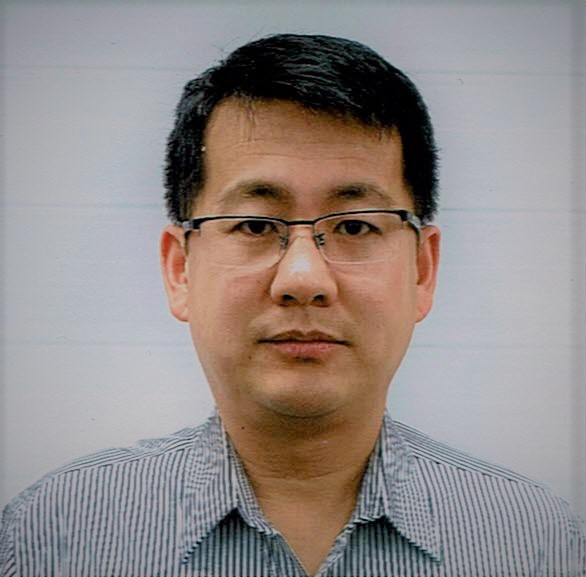}}]{Dr. Naigang Wang}
 received his PhD in materials science and engineering from University of Florida in 2010. Since then, he has been affiliated with IBM T. J. Watson Research Center as a research staff member. His current research focus is on hardware-friendly deep learning algorithms for artificial intelligence accelerators.
\end{IEEEbiography}

\end{document}